\documentclass[twoside,11pt]{article}

\usepackage{jmlr2e}

\usepackage{mypackages}
\usepackage{mycommands}
\hypersetup{ hidelinks }

\makeatletter
\def\cleartheorem#1{\expandafter\let\csname#1\endcsname\relax
    \expandafter\let\csname c@#1\endcsname\relax
}
\makeatother
\cleartheorem{example}
\cleartheorem{theorem}
\cleartheorem{lemma}
\cleartheorem{proposition}
\cleartheorem{definition}
\cleartheorem{corollary}

\newtheorem{theorem}{Theorem}
\newtheorem{lemma}{Lemma}

\newtheorem{definition}{Definition}

\newcommand{\HH} {{\cal H}}

\newcommand{\rb} {{\rangle}}
\newcommand{\lb} {{\langle}}

\newcommand{\E} {{\mathbb E}}

\newcommand{\R} {{\mathbb R}}

\newcommand{\N} {{\mathbb N}}

\usepackage{lastpage}
\jmlrheading{25}{2024}{1-\pageref{LastPage}}{11/23; Revised
10/24}{10/24}{23-1573}{Florentin Guth, Brice Ménard, Gaspar Rochette, and Stéphane Mallat}
\ShortHeadings{A Rainbow in Deep Network Black Boxes}{Guth, Ménard, Rochette, and Mallat}
\firstpageno{1}

\title{A Rainbow in Deep Network Black Boxes}
\author{\name Florentin Guth\thanks{\ Work done while at \'Ecole Normale Sup\'erieure.} \email florentin.guth@nyu.edu \\
       \addr Center for Data Science, New York University, 60 5th Avenue, New York, NY 10011, USA \\  
       Flatiron Institute, 162 5th Avenue, New York, NY 10010, USA
       \AND
       \name Brice Ménard \email menard@jhu.edu \\
       \addr Department of Physics \& Astronomy, Johns Hopkins University \\
       Baltimore, MD 21218, USA
       \AND
       \name Gaspar Rochette \email gaspar.rochette@ens.fr \\
       \addr Département d'informatique, \'Ecole Normale Supérieure, CNRS, PSL University \\
       45 rue d'Ulm, 75005 Paris, France
       \AND
       \name Stéphane Mallat \email stephane.mallat@ens.fr \\
       \addr Collège de France, 11, place Marcelin-Berthelot 75231 Paris, France \\
       Flatiron Institute, 162 5th Avenue, New York, NY 10010, USA
       }
\editor{Lorenzo Rosasco}

\begin{document}

\maketitle

\begin{abstract}%
  A central question in deep learning is to understand the functions learned by deep networks. What is their approximation class? Do the learned weights and representations depend on initialization?
  Previous empirical work has evidenced that kernels defined by network activations are similar across initializations. For shallow networks, this has been theoretically studied with random feature models, but an extension to deep networks has remained elusive.
  Here, we provide a deep extension of such random feature models, which we call the rainbow model. We prove that rainbow networks define deterministic (hierarchical) kernels in the infinite-width limit. The resulting functions thus belong to a data-dependent RKHS which does not depend on the weight randomness.
  We also verify numerically our modeling assumptions on deep CNNs trained on image classification tasks, and show that the trained networks approximately satisfy the rainbow hypothesis. In particular, rainbow networks sampled from the corresponding random feature model achieve similar performance as the trained networks. Our results highlight the central role played by the covariances of network weights at each layer, which are observed to be low-rank as a result of feature learning.
\end{abstract}

\begin{figure*}[t]
  \centering
  \includegraphics[width=\textwidth]{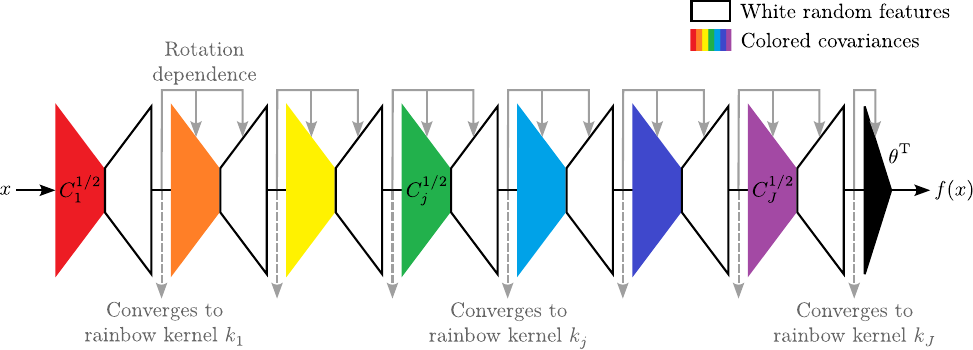}
  \caption{\small
     A deep rainbow network cascades random feature maps whose weight distributions are learned. They typically have a low-rank covariance. Each layer can be factorized into a linear dimensionality reduction determined by the ``colored'' (i.e., non-identity) covariance, followed by a non-linear high-dimensional embedding with ``white'' random features. At each layer, the hidden activations define a kernel which converges to a deterministic rainbow kernel in the infinite-width limit. The activations are however randomly rotated, which induces a similar rotation of the next layer weights. 
     }
  \label{fig:rainbow_model}
\end{figure*}

\section{Introduction}

The weight matrices of deep networks are learned by performing stochastic gradient descent from a random initialization. Each training run thus results in a different set of weights, which can be considered as a random realization of some probability distribution. This randomness is a major challenge in analyzing the learned weights. Finding deterministic quantities which are independent of the details of the initialization and training is thus of great importance to study deep learning.

A prominent example is the kernels defined by hidden activations of wide networks, as has been empirically shown by \citet{raghu2017svcca,kornblith-cka}. That is, if we denote $\hat \phi_j(x)$ and $\hat\phi_j'(x)$ the $j$-th layer feature maps of two wide networks,
\begin{equation}
    \label{eq:empirical_kernel_concentration}
    \innerr{\hat \phi_j(x), \hat\phi_j(x')} \approx \innerr{\hat \phi_j'(x), \hat\phi_j'(x')}, \quad \forall j, x, x'.
\end{equation}
This concentration of kernels has been studied in one-hidden-layer networks in the ``mean-field'' limit.
Under this limit, neuron weights $w_1, \dots, w_{d_1}$ can be modeled as independent samples from a distribution $\pi_1$. The first layer thus computes random features $\hat \phi_1(x) = \frac1{\sqrt{d_1}} (\sigma(\innerr{w_i, x}))_{i \leq d_1}$, whose kernel concentrates as a consequence of the law of large numbers \citep{rahimi-recht-random-features}:
\begin{equation}
    \label{eq:mean_field_kernel_shallow}
    \innerr{\hat \phi_1(x), \hat\phi_1(x')} = \frac1{d_1} \summ i{d_1} \sigma(\innerr{w_i, x}) \, \sigma(\innerr{w_i, x'}) \xrightarrow[d_1 \to \infty]{} \expect[w \sim \pi_1]{\sigma(\innerr{w, x}) \, \sigma(\innerr{w, x'})}.
\end{equation}
Although different networks may have different weights $w'_i \neq w_i$ (even up to permutations), the key point is that they are sampled from the same distribution $\pi_1$. \Cref{eq:empirical_kernel_concentration} is then a consequence of \cref{eq:mean_field_kernel_shallow}. 
However, it is not clear a priori how to extend this analysis beyond the first layer.

In this paper, we introduce a deep extension of the shallow random feature model above to explain \cref{eq:empirical_kernel_concentration} at all layers. It rests on the observation that weight distributions in hidden layers need to be \emph{aligned} based on the previous layer weights.
Indeed, the concentration of kernels is equivalent to the concentration of activations \emph{up to rotations}: a restatement of \cref{eq:empirical_kernel_concentration} is that there exists orthogonal transforms $\hat A_j$ such that
\begin{equation*}
    \hat \phi_j(x) \approx \hat A_j\, \hat\phi_j'(x), \quad \forall j, x.
\end{equation*}
In the one-layer random feature model, $\hat A_1 \neq \Id$ arises from the differences between individual neuron weights $w'_i \neq w_i$.
We empirically observe that weights in hidden layers also rotate together with their input activations: that is, the distribution of weights at layer $j+1$ are the same \emph{after a rotation by $\hat A_{j}$}.
When taken as an assumption, this allows iterating the argument in \cref{eq:mean_field_kernel_shallow} to establish \cref{eq:empirical_kernel_concentration} at all layers.
In summary, kernels concentrate because the underlying weight distributions are the same, and activations rotate because the individual neurons are different samples from these common distributions. 
The rainbow model captures the interaction between these two properties across layers, shedding a theoretical light on the phenomenological \cref{eq:empirical_kernel_concentration}.

The rainbow model is parameterized by random feature distributions for each layer. A special case of interest arises when these distributions are assumed to be Gaussian and centered, so that the model is entirely specified by weight covariance matrices. While the resulting Gaussian rainbow model appears too restrictive to model all trained networks,  we show that it can approximately hold for architectures which incorporate prior information and restrict their learned weights. In some of our numerical experiments, we will thus consider learned scattering networks \citep{separationiclr,phase-collapse}, which have fixed wavelet spatial filters and learn weights along channels only.

Under the Gaussian rainbow model, the weight covariances completely specify the network output. The eigenvectors of these weight covariances can be interpreted as learned features, rather than individual neuron weights which are random. We show numerically that weights of trained networks typically have low-rank covariances. The corresponding rainbow networks thus implement dimensionality reductions in-between the high-dimensional random feature embeddings. 
We further demonstrate that input activation covariances provide efficient approximations of the eigenspaces of the weight covariances. The number of model parameters and hence the supervised learning complexity can thus be considerably reduced by unsupervised information.

This paper makes the following main contributions:
\begin{itemize}
    \item We numerically demonstrate that both hidden activations and weight distributions (specifically, their covariances) of deep networks trained from different initializations concentrate up to rotations when the width increases. This complements previous observations of concentration of the kernels defined by the activations.

    \item 
    We prove that networks sampled from the rainbow model define activations which converge to a random rotation of a deterministic kernel feature vector in the infinite-width limit. In random feature models, the concentration of kernels and activations up to rotations is thus a consequence of the rotation of the weight distributions.

    \item We validate empirically the stronger Gaussian rainbow model for scattering networks trained on CIFAR-10. We verify that the learned weights are approximately Gaussian. Their covariances are sufficient to sample new networks that achieve comparable classification accuracy when the width is large enough.
    Further, we show that SGD training only updates the weight covariances while nearly preserving the white random feature initializations, suggesting a possible dynamical explanation for the Gaussian rainbow assumption in this setting.

    \item We show that the weight covariances of trained deep networks are approximately low-rank, and their dimensionality can be reduced without harming performance by performing PCA on the weights. Further, this can be well-approximated by a PCA on the input activations. In the context of the Gaussian rainbow model, this shows that feature learning amounts to finding an informative subspace, which can be approximated with unsupervised information. More generally, these observations reveal properties of the weight distributions in trained networks that can also be of independent interest.

\end{itemize}

The rainbow model is illustrated in \Cref{fig:rainbow_model}.
In \Cref{sec:model}, we introduce rainbow networks and the associated kernels that describe their infinite-width limit. We validate numerically the above properties and results in \Cref{sec:validation}. Code to reproduce all our experiments can be found at \url{https://github.com/FlorentinGuth/Rainbow}.

\subsection{Related work}

Several strands of research have studied functional properties of neural networks. However, they only apply to networks around their initialization, or to shallow networks. We briefly review these lines of research, and show how our work addresses some of the gaps in the literature.

\paragraph{Lazy versus feature learning.}
For some weight initialization schemes, \citet{jacot-ntk} and \citet{lee-shol-dickstein-ntk} have shown that trained weights have vanishing deviations from their initialization. In these cases, learning is in a ``lazy'' regime \citep{chizat-oyallon-bach-lazy-training} specified by a fixed kernel. It has been opposed to a ``rich'' or feature-learning regime \citep{chizat-bach-implicit-bias-one-hidden-layer,woodworth-soudry-srebro-kernel-vs-rich-regime}, which achieves higher performance on complex tasks \citep{lee-sohl-dickstein-lazy-vs-feature-learning,geiger-wyart-lazy-vs-feature-learning}. While we do not model training dynamics, the rainbow model captures weight distributions that are significantly different from their initialization. These weight distributions depend on the training data, and thus induce a data-dependent kernel, indirectly incorporating feature learning.

\paragraph{Random features and neural network Gaussian processes.}
In a different but related direction, many works have considered networks where all but the last layers are frozen to their initialization, starting from \citep{jarrett2009best,pinto-random-feature-screening}. Such networks compute random features, which specify a kernel that becomes deterministic in the infinite-width limit. This was first studied by \citet{rahimi-recht-random-features} for the one-layer case and generalized to deep architectures by \citet{daniely-singer-random-network-hilbert-spaces}.
As a result, in the infinite-width, networks at initialization represent a function sampled randomly from a Gaussian process \citep{neal-nngp,williams-nngp,nngp-sohl-dickstein,nngp-matthews}. Training is then modeled as performing Bayesian inference by conditioning this prior on the training data.
These approaches thus correspond to performing regression with a fixed kernel, again precluding feature learning (though finite-width corrections can be incorporated, see \citealp{seroussi-ringel-nngp-finite-width-corrections}). Our theoretical results extend this line of work by considering much more general weight distributions, which incorporate dependencies across layers as a result of feature learning.

\paragraph{Mean-field models.}
Feature learning has been precisely studied for one-hidden-layer networks, again in the infinite-width limit \citep{chizat-bach-mean-field,mei-montanari-mean-field,rotskoff-vanden-eijnden-mean-field,sirignano-spiliopoulos-mean-field}. These ``mean-field'' approaches analyze the neuron weight distribution as it evolves away from the normal initialization during training. However, generalizing this result to deeper networks has been challenging due to the dependence between weights across layers \citep{sirignano-spiliopoulos-multilayer-mean-field,e-wojtowytsch-multilayer-mean-field,nguyen-pham-multilayer-mean-field,chen-vanden-eijnden-bruna-three-layer-mean-field,yang-feature-learning-infinite-width}. In this work, we propose a model of such a dependence through the use of representation alignment.

\paragraph{Representation alignment and hierarchical kernels.}
A key observation for generalizing mean-field random feature models to deeper networks was made by \citet{raghu2017svcca,kornblith-cka}. They demonstrated empirically that after an alignment procedure, activations of deep networks trained on the same task from different initializations become increasingly similar as the width increases, at all layers. Equivalently, the activations at a given layer asymptotically define the same deterministic kernel independently of the initialization \citep{kriegeskorte-rsa,williams-kornblith-shape-metrics}.
The network thus belongs to a hierarchical reproducing kernel Hilbert space similar to the ones studied by \citet{cho-saul,anselmi-rosasco-poggio-hierarchical-kernels-groups,mairal-ckn-learning-projections,BiettiThese}. However, characterizing these kernels requires understanding how they depend on the data distribution in order to capture feature learning. This was done for the one-layer case by \citet{pandey-harris-colored-random-features-sensory-encoding} by considering structured (non-isotropic) random feature kernels. \cite{bordelon-pehlevan-dynamical-mean-field} characterized the evolution of activation and gradient kernels in deep networks during training with self-consistent equations from dynamical mean-field theory, which are often challenging to solve. Our rainbow model builds on these ideas to give an integrated picture of the approximation class of deep neural networks.

\section{Rainbow networks}
\label{sec:model}

Weight matrices of learned deep networks are strongly dependent across layers. Deep rainbow networks define a mathematical model of these dependencies through rotation matrices that align input activations at each layer. We review in \Cref{sec:random_features_alignment} the properties of random features, which are the building blocks of the model. We then introduce in \Cref{sec:infinite_rainbow_kernel} deep fully-connected rainbow networks, which cascade aligned random feature maps. We show in \Cref{sec:prior_convolutional_architectures} how to incorporate inductive biases in the form of symmetries or local neuron receptive fields. We also extend rainbow models to convolutional networks.

\subsection{Rotations in random feature maps}
\label{sec:random_features_alignment}

We being by reviewing the properties of one-hidden layer random feature networks. We then prove that random weight fluctuations produce a random rotation of the hidden activations in the limit of infinite width (in a sense made precise in \Cref{th:convergence_shallow} below). The rainbow model will allow us to apply this result at all layers of a deep network in \Cref{sec:infinite_rainbow_kernel}.

\paragraph{Random feature network.}
A one-hidden layer network computes a hidden activation layer with a matrix $W$ of size $d_1 \times d_0$ and a pointwise non-linearity $\sigma$:
\begin{equation*}
\hat \varphi(x) =  \sigma (W x) ~~\mbox{for}~~x \in \R^{d_0} .
\end{equation*}
We consider a random feature network \citep{rahimi-recht-random-features}. The rows of $W$, which contain the weights of different neurons, are independent and have the same probability distribution $\pi$:
\begin{equation*}
    W = (w_i )_{i \leq d_1}~~\mbox{with i.i.d.}~~w_i \sim \pi .
\end{equation*}
In many random feature models, each row vector has a known distribution with uncorrelated coefficients \citep{jarrett2009best,pinto-random-feature-screening}. Learning is then reduced to calculating the output weights $\hat \theta$, which define
\begin{equation*}
\hat f(x) = \lb \hat \theta ,  \hat \varphi (x) \rb .
\end{equation*}
In contrast, we consider general distributions $\pi$ which will be estimated from the weights of trained networks in \Cref{sec:validation}. Considering more general random feature distributions with non-identity covariance has been shown to greatly improve modeling of sensory neuron receptive fields \citep{pandey-harris-colored-random-features-sensory-encoding}.

Our network does not include any bias for simplicity. Bias-free networks have been shown to achieve comparable performance as networks with biases for denoising \citep{mohan-kadkhodaie-simoncelli-bias-free-denoising} and image classification \citep{separationiclr,phase-collapse}. However, biases can easily be incorporated in random feature models and thus rainbow networks.

We consider a normalized network, where $\sigma$ includes a division by $\sqrt{d_1}$ so that $\norm{\hat\varphi(x)}$ remains of the order of  unity when the width $d_1$ increases.
We shall leave this normalization implicit to simplify notations, except when illustrating mathematical convergence results. Note that this choice differs from the so-called standard parameterization \citep{yang-feature-learning-infinite-width}. In numerical experiments, we perform SGD training with this standard parameterization which avoids getting trapped in the lazy training regime \citep{chizat-oyallon-bach-lazy-training}. Our normalization convention is only applied at the end of training, where the additional factor of $\sqrt{d_1}$ is absorbed in the next-layer weights $\hat\theta$.

We require that the input data has finite energy: $\expectt[x]{\norm{x}^2} < +\infty$. We further assume that the non-linearity $\sigma$ is Lipschitz continuous, which is verified by many non-linearities used in practice, including ReLU. Finally, we require that the random feature distribution $\pi$ has finite fourth-order moments.

\paragraph{Kernel convergence.}
We now review the convergence properties of one-hidden layer random feature networks. This convergence is captured by the convergence of their kernel \citep{rahimi-recht-random-features,rahimi-recht-uniform-approximation},
\begin{equation*}
    \hat k (x, x') =  \innerr{\hat \varphi(x),
    \hat \varphi(x')} = \frac1{d_1} \sum_{i=1}^{d_1} \sigma\paren{\inner{w_i, x}} \, \sigma\parenn{\innerr{w_i, x'}},
\end{equation*}
where we have made explicit the factor $d_1\pinv$ coming from our choice of normalization. Since the rows $w_i$ are independent and identically distributed, the law of large numbers implies that when the width $d_1$ goes to infinity, this empirical kernel has a mean-square convergence to the asymptotic kernel
\begin{equation}
    \label{eq:asymptotic_kernel}
    k(x,x') = \expect[w\sim \pi]{ \sigma\paren{\inner{w, x}} \, \sigma\parenn{\innerr{w, x'}} }.
\end{equation}
This convergence means that even though $\hat \varphi$ is random, its kernel is asymptotically deterministic. As we will see, this imposes that random fluctuations of $\hat\varphi(x)$ are reduced to rotations in the large $d_1$ limit.

Let $\varphi(x)$ be an infinite-dimensional deterministic feature vector in a separable Hilbert space $H$, which satisfies
\begin{equation}
    \label{eq:varphi_def}
    \innerr{\varphi(x), \varphi(x')}_H = k(x,x').
\end{equation}
Such feature vectors always exist (\citealp{aronszajn-theory-reproducing-kernels}, see also \citealp{scholkopf-smola-kernel-book}). For instance, one can choose $\varphi(x) = \paren{\sigma(\inner{w, x})}_{w}$, the infinite-width limit of random features $\sigma W$. In that case, $H = L^2(\pi)$, that is, the space of square-integrable functions with respect to $\pi$, with dot-product $\inner{g,h}_H = \expectt[w\sim\pi]{g(w)\,h(w)}$. This choice is however not unique: one can obtain other feature vectors defined in other Hilbert spaces by applying a unitary transformation to $\varphi$, which does not modify the dot product in \cref{eq:varphi_def}. In the following, we choose the kernel PCA (KPCA) feature vector, whose covariance matrix $\expectt[x]{\varphi(x)\, \varphi(x)\trans}$ is diagonal with decreasing values along the diagonal, introduced by \citet{scholkopf-kpca}. It is obtained by expressing any feature vector $\varphi$ in its PCA basis relative to the distribution of $x$. In this case $H = \ell^2(\N)$.

Finally, we denote by $\cal H$ the reproducing kernel Hilbert space (RKHS) associated to the kernel $k$ in \cref{eq:asymptotic_kernel}. It is the space of functions $f$ which can be written $f(x) = \inner{\theta, \varphi(x)}_H$, with norm $\norm{f}_{\hilbert} = \norm{\theta}_H$.\footnote{We shall always assume that $\theta$ is the minimum-norm vector such that $f(x) = \inner{\theta, \varphi(x)}_H$.} A random feature network defines approximations of functions in this RKHS. With $H = L^2(\pi)$, these functions can be written
\begin{equation*}
    f(x) = \expect[w \sim \pi]{\theta(w) \, \sigma(\inner{w, x})} = \int \theta(w) \, \sigma(\inner{w, x}) \,\diff \pi(w).
\end{equation*}
This expression is equivalent to the mean-field limit of one-hidden-layer networks \citep{chizat-bach-mean-field,mei-montanari-mean-field,rotskoff-vanden-eijnden-mean-field,sirignano-spiliopoulos-mean-field}, which we will generalize to deep networks in \Cref{sec:infinite_rainbow_kernel}.

\paragraph{Rotational alignment.}
We now introduce rotations which align approximate kernel feature vectors. By abuse of language, we use rotations as a synonym for orthogonal transformations, and also include improper rotations which are the composition of a rotation with a reflection. Here, we prove that \emph{aligned} features vectors $\hat \varphi$ converge to their infinite-width counterpart $\varphi$.

We begin by an informal derivation of our main result.
We have seen that the kernel $\hat k(x,x') = \lb \hat \varphi(x) , \hat \varphi(x') \rb$ converges to the kernel $k(x,x') = \lb \varphi(x) , \varphi(x') \rb$.  We thus expect, and will later prove, that for large widths there exists a rotation $\hat A$ such that $\hat A \, \hat \varphi \approx \varphi$ (in a sense made precise by \Cref{th:convergence_shallow} below), because all feature vectors of the kernel $k$ are rotations of one another. The rotation $\hat A$ is dependent on the random feature realization $W$ and is thus random. The network activations $\hat\varphi(x) \approx \hat A\trans \varphi(x)$ are therefore a random rotation of the deterministic feature vector $\varphi(x)$. For the KPCA feature vector $\varphi$, $\hat A$ approximately computes an orthonormal change of coordinate of $\hat\varphi(x)$ to its PCA basis.

For any function $f(x) = \inner{\theta, \varphi(x)}_H$ in $\hilbert$, if the output layer weights of the network are $\hat\theta = \hat A\trans \theta$, then the network output is
\begin{equation*}
    \hat f(x)  = \innerr{\hat A\trans \theta, \hat\varphi(x)} = \innerr{\theta, \hat A\, \hat\varphi(x)}_H \approx f(x).
\end{equation*}
This means that the final layer coefficients $\hat\theta$ can cancel the random rotation $\hat A$ introduced by $W$, so that the random network output $\hat f(x)$ converges when the width $d_1$ increases to a fixed function in $\cal H$. This propagation of rotations across layers is key to understanding the weight dependencies in deep networks. We now make the above arguments more rigorous and prove that $\hat \varphi$ and $\hat f$ respectively converge to $\varphi$ and $f$, for an appropriate choice of $\hat A$.

We write ${\cal O}(d_1)$ the set of linear operators $\hat A$ from $\R^{d_1}$ to $H = \ell^2(\NN)$ which satisfy $\hat A\trans \hat A = \Id_{d_1}$. Each $\hat A \in \mathcal{O}(d_1)$ computes an isometric embedding of $\RR^{d_1}$ into $H$, while $\hat A\trans$ is an orthogonal projection onto a $d_1$-dimensional subspace of $H$ which can be identified with $\RR^{d_1}$. The alignment $\hat A$ of $\hat \varphi$ to $\varphi$ is defined as the minimizer of the mean squared error:
\begin{equation}
    \label{eq:alignment_objective}
    \hat A = \argmin_{\hat A \in {\cal O}(d_1)}\
    \expect[x]{\normm{\hat A\,\hat\varphi(x) - \varphi(x)}^2_H}.
\end{equation}
This optimization problem, known as the (orthogonal) Procrustes problem \citep{hurley-cattel-procrustes-problem,schonemann-procrustes-solution}, admits a closed-form solution, computed from a singular value decomposition of the (uncentered) cross-covariance matrix between $\varphi$ and $\hat\varphi$:
\begin{equation}
    \label{eq:alignment_closed_form}
    \hat A = UV\trans ~~\text{with}~~ \expect[x]{\varphi(x) \, \hat\varphi(x)\trans} = USV\trans.
\end{equation}
The mean squared error (\ref{eq:alignment_objective}) of the optimal $\hat A$ (\ref{eq:alignment_closed_form}) is then
\begin{equation}
    \label{eq:similarity_measure}
    \expect[x]{\normm{\hat A\,\hat\varphi(x) - \varphi(x)}^2_H} = \tr{\expect[x]{\hat\varphi(x)\,\hat\varphi(x)\trans}} + \tr{\expect[x]{\varphi(x)\,\varphi(x)\trans}} - 2\norm{\expect[x]{\varphi(x) \, \hat\varphi(x)\trans}}_1,
\end{equation}
where $\norm{\cdot}_1$ is the nuclear (or trace) norm, that is, the sum of the singular values. \Cref{eq:similarity_measure} defines a distance between the representations $\hat \varphi$ and $\varphi$ which is related to various similarity measures used in the literature.\footnote{By normalizing the variance of $\varphi$ and $\hat\varphi$, \cref{eq:similarity_measure} can be turned into a similarity measure $\normm{\expectt[x]{\varphi(x) \, \hat\varphi(x)\trans}}{}_1 / \sqrt{\expectt[x]{\normm{\varphi(x)}^2} \expectt[x]{\normm{\hat \varphi(x)}^2}}$.
It is related to the kernel alignment used by \citet{cristianini-kernel-alignment,cortes-cka,kornblith-cka}, although the latter is based on the Frobenius norm of the cross-covariance matrix $\expectt[x]{\varphi(x) \, \hat\varphi(x)\trans}$ rather than the nuclear norm. Both similarity measures are invariant to rotations of either $\varphi$ or $\hat\varphi$ and therefore only depend on the kernels $k$ and $\hat k$, but the nuclear norm has a geometrical interpretation in terms of an explicit alignment rotation (\ref{eq:alignment_closed_form}). Further, \Cref{app:proof-convergence-shallow} shows that the formulation (\ref{eq:similarity_measure}) has connections to optimal transport through the Bures-Wasserstein distance \citep{bhatia-bures-wasserstein}.
Canonical correlation analysis also provides an alignment, although not in the form of a rotation. It is based on a singular value decomposition of the cross-correlation matrix $\expectt[x]{\varphi(x) \, \varphi(x)\trans}\prsqrtt \, \expectt[x]{\varphi(x) \, \hat\varphi(x)\trans} \, \expectt[x]{\hat\varphi(x) \, \hat\varphi(x)\trans}\prsqrtt$ rather than the cross-covariance, and is thus sensitive to noise in the estimation of the covariance matrices \citep{raghu2017svcca,morcos-raghu-bengio-proj-cca}. Equivalently, it corresponds to replacing $\varphi$ and $\hat\varphi$ with their whitened counterparts $\expectt[x]{\varphi(x) \, \varphi(x)\trans}\prsqrtt \, \varphi$ and $\expectt[x]{\hat\varphi(x) \, \hat\varphi(x)\trans}\prsqrtt \, \hat\varphi$ in \cref{eq:alignment_objective,eq:alignment_closed_form,eq:similarity_measure}.
}

The alignment rotation (\ref{eq:alignment_objective},\ref{eq:alignment_closed_form}) was used by \citet{haxby-hyperalignment} to align fMRI response patterns of human visual cortex from different individuals, and by \citet{smith-word-embedding-procrustes} to align word embeddings from different languages. Alignment between network weights has also been considered in previous works, but it was restricted to permutation matrices \citep{entezari-permutation-conjecture,permutation-connectivity,git-rebasin}. Permutations have the advantage of commuting with pointwise non-linearities, and can therefore be introduced while exactly preserving the network output function. However, they are not sufficiently rich to capture the variability of random features. It is shown in \citet{entezari-permutation-conjecture} that the error after permutation alignment converges to zero with the number of random features $d_1$ at a polynomial rate which is cursed by the dimension $d_0$ of $x$. On the contrary, the following theorem proves that the error after rotational alignment has a convergence rate which is independent of the dimension $d_0$.

\begin{theorem}
\label{th:convergence_shallow}
Assume that $\expectt[x]{\normm{x}^2} < +\infty$, $\sigma$ is Lipschitz continuous, and $\pi$ has finite fourth order moments. Then there exists a constant $c > 0$ which does not depend on $d_0$ nor $d_1$ such that
\begin{equation*}
    \expect[W,x,x']{\abss{\hat k(x,x') - k(x,x')}{}^2} \leq c\, d_1^{-1} ,
\end{equation*}
where $x'$ is an i.i.d.\ copy of $x$. Suppose that the sorted eigenvalues $\lambda_{1} \geq \cdots \geq \lambda_m \geq \cdots$ of $\E_x[\varphi(x) \, \varphi(x)\trans]$ satisfy $\lambda_m = O(m^{-\alpha})$ with $\alpha > 1$. Then the alignment $\hat A$ defined in (\ref{eq:alignment_objective}) satisfies
\begin{equation*}
{\expect[W,x]{ \normm{\hat A \, \hat \varphi(x) - \varphi(x)}^2_H }} \leq c \, d_1^{-\eta}
~~\mbox{with}~~\eta = \frac{\alpha - 1}{2(2\alpha - 1)} > 0.
\end{equation*}
Finally, for any $f(x) = \inner{\theta, \varphi(x)}_H$ in ${\cal H}$, if $\hat \theta = \hat A \trans \theta$ then
\begin{equation*}
  {\expect[W,x]{|\hat f(x) - f (x)|^2}} \leq c \, \norm{f}^2_\hilbert \,  d_1^{-\eta}.
\end{equation*}
\end{theorem}

The proof is given in \Cref{app:proof-convergence-shallow}. The convergence of the empirical kernel $\hat k$ to the asymptotic kernel $k$ is a direct application of the law of large numbers. The mean-square distance (\ref{eq:similarity_measure}) between $\hat A \, \hat\varphi$ and $\varphi$ is then rewritten as the Bures-Wasserstein distance \citep{bhatia-bures-wasserstein} between the kernel integral operators associated to $\hat k$ and $k$. It is controlled by their mean-square distance via an entropic regularization of the underlying optimal transport problem (\citealp{cuturi-entropic-optimal-transport}, see also \citealp{peyre-cuturi-optimal-transport-book}). The convergence rate is then obtained by exploiting the eigenvalue decay of the kernel integral operator.

\Cref{th:convergence_shallow} proves that there exists a rotation $\hat A$ which nearly aligns the hidden layer of a random feature network with any feature vector of the asymptotic kernel, with an error which converges to zero. The network output converges if that same rotation is applied on the last layer weights. We will use this result in the next section to define deep rainbow networks, but we note that it can be of independent interest in the analysis of random feature representations.
The theorem assumes a power-law decay of the covariance spectrum of the feature vector $\varphi$ (which is independent of the choice of $\varphi$ satisfying eq.~\ref{eq:varphi_def}). Because $\summ m\infty \lambda_m = \expectt[x]{\norm{\varphi(x)}^2} < +\infty$ (as shown in the proof), a standard result implies that $\lambda_m = o(m^{-1})$, so the assumption $\alpha > 1$ is not too restrictive.
The constant $c$ is explicit and depends polynomially on the constants involved in the hypotheses (except for the exponent $\alpha$).
The convergence rate $\eta = \frac{\alpha - 1}{2(2\alpha - 1)}$ is an increasing function of the power-law exponent $\alpha$. It vanishes in the critical regime when $\alpha \to 1$, and increases to $\frac14$ when $\alpha \to \infty$.
This bound might be pessimistic in practice, as a heuristic argument suggests a rate of $\frac12$ when $\alpha \to \infty$ based on the rate $1$ on the kernels. A comparison with convergence rates of random features KPCA \citep{sriperumbudur-kernel-pca-rf} indeed suggests it might be possible to improve the convergence rate to $\frac{\alpha - 1}{2\alpha - 1}$.
Finally, although we give results in expectation for the sake of simplicity, we note that bounds in probability can be obtained using Bernstein concentration bounds for operators \citep{tropp-bernstein,minsker-bernstein} in the spirit of \citet{rudi-rosasco-subspace-learning,bach-random-features}.

\subsection{Deep rainbow networks}
\label{sec:infinite_rainbow_kernel}

The previous section showed that the hidden layer of a random feature network converges to an infinite-dimensional feature vector, up to a rotation defined by the alignment $\hat A$. This section defines deep fully-connected rainbow networks by cascading conditional random features, whose kernels also converge in the infinite-width limit. It provides a model of the joint probability distribution of weights of trained networks, whose layer dependencies are captured by alignment rotation matrices.

We consider a deep fully-connected neural network with $J$ hidden layers, which iteratively transforms the input data $x \in \RR^{d_0}$ with weight matrices $W_j$ of size $d_{j} \times d_{j-1}$ and a pointwise non-linearity $\sigma$, to compute each activation layer of depth $j$:
\begin{equation*}
    \hat \phi_j (x) =   \sigma W_j\, \cdots \, \sigma W_1 \, x .
\end{equation*}
$\sigma$ includes a division by $\sqrt{d_j}$, which we do not write explicitly to simplify notations. After $J$ non-linearities, the last layer outputs
\[
\hat f (x) = \lb \hat \theta , \hat \phi_J (x) \rb .
\]

\paragraph{Infinite-width rainbow networks.}
The rainbow model defines each $W_{j}$ conditionally on the previous $(W_{\ell})_{\ell < j}$ as a random feature matrix. The distribution of random features at layer $j$ is rotated to account for the random rotation introduced by $\hat\phi_{j-1}$. We first introduce infinite-width rainbow networks which define the asymptotic feature vectors used to compute these rotations.

\begin{definition}
\label{def:infinite_rainbow_network}
An infinite-width rainbow network has activation layers defined in a separable Hilbert space $H_j$ for any $j \leq J$ by
\begin{equation*}
    \phi_j (x) = \varphi_j (\varphi_{j-1}(\dots \varphi_1(x) \dots)) \in H_j ~~\mbox{for}~~x \in H_0 = \R^{d_0} ,
\end{equation*}
where each $\varphi_j \colon H_{j-1} \to H_j$ is defined from a probability distribution $\pi_j$ on $H_{j-1}$ by
\begin{equation}
    \label{eq:feature_vector_def_deep}
    \lb \varphi_j(z) , \varphi_j (z') \rb_{H_j}  = \expect[w\sim \pi_j]{ \sigma\parenn{\inner{w, z}_{H_{j-1}}} \, \sigma\parenn{\innerr{w, z'}_{H_{j-1}}} } ~~\mbox{for}~~ z,z' \in H_{j-1} .
\end{equation}
It defines a rainbow kernel
\[
k_j (x,x') = \lb \phi_j(x) , \phi_j(x') \rb_{H_j}.
\]
For $\theta \in H_J$, the infinite-width rainbow network outputs
\[
f(x) = \lb \theta , \phi_J (x) \rb_{H_J}  \in \HH_J,
\]
where $\HH_J$ is the RKHS of the rainbow kernel $k_J$ of the last layer.
If all probability distributions $\pi_j$ are Gaussian, then the rainbow network is said to be Gaussian.
\end{definition}

Each activation layer $\phi_j(x) \in H_j$ of an infinite-width rainbow network has an infinite dimension and is deterministic. We shall see that the cascaded feature maps $\varphi_j$ are infinite-width limits of $\sigma W_j$ up to rotations.
One can arbitrarily rotate a feature vector $\varphi_j(z)$ which satisfies (\ref{eq:feature_vector_def_deep}), which also rotates the Hilbert space $H_j$ and $\phi_j(x)$. If the distribution $\pi_{j+1}$ at the next layer (or the weight vector $\theta$ if $j =J$) is similarly rotated, this operation preserves the dot products $\innerr{w, \phi_j(x)}_{H_j}$ for $w \sim \pi_{j+1}$. It therefore does not affect the asymptotic rainbow kernels at each depth $j$:
\begin{equation}
    \label{eq:rainbow_kernel_depth_j}
    k_j(x,x') =  \expect[w\sim \pi_j]{ \sigma\parenn{\inner{w, \phi_{j-1}(x)}_{H_{j-1}}} \, \sigma\parenn{\innerr{w, \phi_{j-1}(x')}_{H_{j-1}}} } ,
\end{equation}
as well as the rainbow network output $f(x)$. We shall fix these rotations by choosing KPCA feature vectors. This imposes that $H_j = \ell^2(\NN)$ and $\expectt[x]{\phi_j(x)\,\phi_j(x)\trans}$ is diagonal with decreasing values along the diagonal. The random feature distributions $\pi_j$ are thus defined with respect to the PCA basis of $\phi_j(x)$. Infinite-width rainbow networks are then uniquely determined by the distributions $\pi_j$ and the last-layer weights $\theta$.

The weight distributions $\pi_j$ for $j \geq 2$ are defined in the infinite-dimensional space $H_{j-1}$ and some care must be taken. We say that a distribution $\pi$ on a Hilbert space $H$ has bounded second-order moments if its (uncentered) covariance operator $\expectt[w\sim\pi]{ww\trans}$ is bounded (for the operator norm). The expectation is to be understood in a weak sense: we assume that there exists a bounded operator $C$ on $H$ such that $z\trans C z' = \expectt[w\sim\pi]{\innerr{w, z}_H\innerr{w, z'}_H}$ for $z,z' \in H$. We further say that $\pi$ has bounded fourth-order moments if for every trace-class operator $T$ (that is, such that $\tr (T\trans T)\psqrtt  < + \infty$), $\expectt[w\sim \pi]{\parenn{w\trans T w}^2} < +\infty$. We will assume that the weight distributions $\pi_j$ have bounded second-{} and fourth-order moments. Together with our assumptions that $\expectt[x]{\normm{x}^2} < +\infty$ and that $\sigma$ is Lipschitz continuous, this verifies the existence of all the infinite-dimensional objects we will use in the sequel. For the sake of brevity, we shall not mention these verifications in the main text and defer them to \Cref{app:proof-convergence-deep}. Finally, we note that we can generalize rainbow networks to cylindrical measures $\pi_j$, which define cylindrical random variables $w$ (\citealp{vakhania-probability-banach}, see also \citealp{riedle-cylindrical-wiener-process} or \citealp[][Section 2.1.1]{gawarecki-mandrekar-sde-infinite-dim}). Such cylindrical random variables $w$ are linear maps such that $w(z)$ is a real random variable for every $z \in H_{j-1}$. $w(z)$ cannot necessarily be written $\inner{w, z}$ with a random $w \in H_{j-1}$. We still write $\inner{w, z}$ by abuse of notation, with the understanding that it refers to $w(z)$. For example, we will see that finite-width networks at initialization converge to infinite-width rainbow networks with $\pi_j = \normal(0, \Id)$, which is a cylindrical measure but not a measure when $H_{j-1}$ is infinite-dimensional.

\paragraph{Dimensionality reduction.}
Empirical observations of trained deep networks show that they have approximately low-rank weight matrices \citep{martin-mahoney-jmlr,thamm-staats-rosenow-rmt-weights-spectra}. They compute a dimensionality reduction of their input, which is characterized by the singular values of the layer weight $W_j$, or equivalently the eigenvalues of the empirical weight covariance $d_j\pinv\, W_j\trans W_j$. For rainbow networks, the uncentered covariances $C_j = \expectt[w\sim\pi_j]{ww\trans}$ of the weight distributions $\pi_j$ therefore capture the linear dimensionality reductions of the network. If $C^{1/2}_j$ is the symmetric square root of $C_j$, we can rewrite (\ref{eq:feature_vector_def_deep}) with a change of variable as
\begin{align*}
    \varphi_j(z) = \tilde\varphi_j\paren{C_j\psqrtt z} ~~\mbox{with}~~
     \innerr{\tilde\varphi_j(z), \tilde\varphi_j(z')}_{H_j} = \expect[w \sim \tilde\pi_j]{\sigma\parenn{\innerr{w, z}} \, \sigma\parenn{\innerr{z, z'}}},
\end{align*}
where $\tilde\pi_j$ has an identity covariance. Rainbow network activations can thus be written:
\begin{equation}
    \phi_j(x) = \tilde\varphi_j\paren{C_j\psqrtt \cdots \tilde\varphi_1\parenn{C_1\psqrtt x}}.
\end{equation}
Each square root $C_j\psqrtt$ performs a linear dimensionality reduction of its input, while the white random feature maps $\tilde\varphi_j$ compute high-dimensional non-linear embeddings. Such linear dimensionality reductions in-between kernel feature maps had been previously considered in previous works \citep{cho-saul,mairal-ckn-learning-projections,BiettiThese}.

\paragraph{Gaussian rainbow networks.}
The distributions $\pi_j$ are entirely specified by their covariance $C_j$ for Gaussian rainbow networks, where we then have
\begin{equation*}
\pi_j = {\cal N}(0,C_j) .
\end{equation*}
When the covariance $C_j$ is not trace-class, $\pi_j$ is a cylindrical measure as explained above. If $\sigma$ is a homogeneous non-linearity such as ReLU, on can derive \citep{cho-saul} from (\ref{eq:rainbow_kernel_depth_j}) that Gaussian rainbow kernels can be written from a homogeneous dot-product:
\begin{equation}
    \label{eq:dot_product_kernel_depth_j}
    k_j(x,x') =  {\|z_j(x)\|\, \|z_j(x')\|}\, \kappa \paren{ \frac{\lb  z_j(x) , z_j(x')  \rb}{\|z_j(x)\|\, \|z_j(x')\|} } ~~\mbox{with}~~ z_j(x) = C_j^{1/2} \phi_{j-1}(x),
\end{equation}
where $\kappa$ is a scalar function which depends on the non-linearity $\sigma$. The Gaussian rainbow kernels $k_j$ and the rainbow RKHS $\HH_J$ only depend on the covariances $(C_j)_{j \leq J}$. If $C_j = \Id$ for each $j$, then $k_j$ remains a dot-product kernel because $\innerr{z_j(x),z_j(x')} = \innerr{\phi_{j-1}(x),\phi_{j-1}(x')} = k_{j-1}(x,x')$. If the norms $\norm{z_j(x)}$ concentrate, we then obtain $k_j(x,x') = \kappa(\dots \kappa(\innerr{x, x'})\dots)$ \citep{daniely-singer-random-network-hilbert-spaces}. Increasing depth then does not lead to better approximation properties\footnote{Though depth may still affects the generalization properties by changing the spectral bias of the kernels \citep{bordelon2020spectrum}.}, as $k_j$ has the same expressivity as $k_1$ \citep{bietti-bach-dot-product-kernel-cascade}. When $C_j \neq \Id$, Gaussian rainbow kernels $k_j$ cannot be written as a cascade of elementary kernels, but their square roots $\phi_j$ are a cascade of kernel feature maps $\varphi_\ell = \tilde\varphi_\ell \, C\psqrtt_\ell$ for $\ell \leq j$. The white random feature maps $\tilde\varphi_j$ have simple expressions as they arise from the homogeneous dot-product kernel:
\begin{align*}
 \innerr{\tilde\varphi_j(z), \tilde\varphi_j(z')}_{H_j} =  {\|z\|\, \|z'\|}\, \kappa \paren{ \frac{\lb  z , z'  \rb} {\|z\| \, \|z'\| } }.
\end{align*}
This dot-product kernel implies that $\tilde\varphi_j$ is equivariant to rotations, and hence symmetry properties on the network $\phi_j$ as we will see in \Cref{sec:prior_convolutional_architectures}.

\paragraph{Finite-width rainbow networks.}
We now go back to the general case of arbitrary weight distributions $\pi_j$ and introduce finite-width rainbow networks, which are random approximations of infinite-width rainbow networks. Each weight matrix $W_{j}$ is iteratively defined conditionally on the previous weight matrices $(W_{\ell})_{\ell < j}$. Its conditional probability distribution is defined in order to preserve the key induction property of the rainbow convergence of the activations $\hat\phi_j$. Informally, it states that $\hat A_{j} \, \hat\phi_{j} \approx \phi_{j}$ where $\hat A_{j} \colon \RR^{d_{j}} \to H_{j}$ is an alignment rotation. Finite-width rainbow networks impose sufficient conditions to obtain this convergence at all layers, as we will show below.

The first layer $W_1$ is defined as in \Cref{sec:random_features_alignment}. Suppose that $W_1,\, \dots, W_{j-1}$ have been defined. By induction, there exists an alignment rotation $\hat A_{j-1} \colon \RR^{d_{j-1}} \to H_{j-1}$, defined by
\begin{equation}
    \label{eq:alignment_objective2}
    \hat A_{j-1} = \argmin_{\hat A \in {\cal O}(d_{j-1})}\
    \expect[x]{\normm{\hat A\,\hat\phi_{j-1}(x) - \phi_{j-1}(x)}^2_{H_{j-1}}},
\end{equation}
such that $\hat A_{j-1} \, \hat\phi_{j-1}(x) \approx \phi_{j-1}(x)$. We wish to define $W_j$ so that $\hat A_j \, \hat \phi_j(x) \approx \phi_j(x)$. This can be achieved with a random feature approximation of $\varphi_{j}$ composed with the alignment $\hat A_{j-1}$. Consider a (semi-infinite) random matrix $W_{j}'$ of $d_{j}$ i.i.d.\ rows in $H_{j-1}$ distributed according to $\pi_j$:
\begin{align*}
W'_{j} &= (w'_{ji})_{i \leq d_{j}}~~\mbox{with i.i.d.}~~w'_{ji} \sim \pi_j .
\end{align*}
We then have $\hat A_j \, \sigma \parenn{W'_j x} \approx \varphi_j(x)$ for a suitably defined $\hat A_j$, as in \Cref{sec:random_features_alignment}. Combining the two approximations, we obtain
\begin{equation*}
    \hat A_j \, \sigma\paren{ W'_j \, \hat A_{j-1} \, \hat\phi_{j-1}(x)} \approx \varphi_j \paren{\phi_{j-1}(x)} = \phi_j(x).
\end{equation*}
We thus define the weight at layer $j$ with the aligned random features
\begin{align*}
    W_j = W'_j \, \hat A_{j-1} .
\end{align*}
It is a random weight matrix of size $d_j \times d_{j-1}$, with rotated rows $\hat A_{j-1}\trans w'_{ji}$ that are independent and identically distributed when conditioned on the previous layers $(W_\ell)_{\ell < j}$. This inverse rotation of random weights cancels the rotation introduced by the random features at the previous layer, and implies a convergence of the random features cascade as we will prove below. This qualitative derivation motivates the following definition of finite-width rainbow networks.

\begin{definition}
\label{def:finite-rainbow-network}
A finite-width rainbow network approximation of an infinite-width rainbow network with weight distributions $(\pi_j)_{j \leq J}$ is defined for each $j \leq J$ by a random weight matrix $W_j$ of size $d_j \times d_{j-1}$ which satisfies
\begin{align}
\label{eq:definition_wj}
    W_j &= (\hat A\trans_{j-1} w'_{ji} )_{i \leq d_j}~~\mbox{with i.i.d.}~~w'_{ji} \sim \pi_j,
\end{align}
where $\hat A_{j-1}$ is the rotation defined in (\ref{eq:alignment_objective2}). The last layer weight vector is $\hat \theta = \hat A_J\trans \theta$ where $\theta$ is the last layer weight of the infinite-width rainbow network.
\end{definition}

The random weights $W_j$ of a finite rainbow networks are defined as rotations and finite-dimensional projections of the $d_j$ infinite-dimensional random vectors $w'_{ji}$, which are independent. The dependence on the previous layers $(W_{\ell})_{\ell < j}$ is captured by the rotation $\hat A_{j-1}$. The rows of $W_j$ are thus not independent, but they are independent when conditioned on $(W_\ell)_{\ell < j}$.

The rotation and projection of the random weights (\ref{eq:definition_wj}) implies a similar rotation and projection on the moments of $W_j$ conditionally on $(W_\ell)_{\ell < j}$. In particular, the conditional covariance of $W_j$ is thus
\begin{equation}
    \label{eq:conditional_covariance}
    \hat C_j = \hat A_{j-1}\trans C_j \hat A_{j-1}.
\end{equation}
$W_j$ can then be factorized as the product of a white random feature matrix $\tilde W_j$ with the covariance square root:
\begin{equation*}
    W_j =  \tilde W_j\,\hat C_j^{1/2}~~\mbox{with i.i.d.}~~\tilde w_{ji}
    ~\mbox{conditionally on}~(W_\ell)_{\ell < j}.
\end{equation*}
Note that the distribution of the white random features $\tilde w_{ji}$ depends in general on $\hat A_{j-1}$. However, for Gaussian rainbow networks with $\pi_j = {\cal N}(0, C_j)$, this dependence is limited to the covariance $\hat C_j$ and $\tilde W_j = G_j$ is a Gaussian white matrix with i.i.d.\ normal entries that are independent of the previous layer weights $(W_\ell)_{\ell < j}$:
\begin{equation}
    \label{eq:gaussian_wj_factorization}
    W_j =  G_j\,\hat C_j^{1/2}~~\mbox{with i.i.d.}~~G_{jik} \sim {\cal N}(0,1)~.
\end{equation}

Finite-width Gaussian rainbow networks are approximation models of deep networks that have been trained end-to-end by SGD on a supervised task. We will explain in \Cref{sec:validation} how each covariance $C_j$ of the rainbow model can be estimated from the weights of one or several trained networks. The precision of a Gaussian rainbow model is evaluated by sampling new weights according to (\ref{eq:gaussian_wj_factorization}) and verifying that the resulting rainbow network has a similar performance as the original trained networks.

\paragraph{Convergence to infinite-width networks.}
The heuristic derivation used to motivate \Cref{def:finite-rainbow-network} suggests that the weights rotation (\ref{eq:definition_wj}) guarantees the convergence of finite-width rainbow networks towards their infinite-width counterpart. This is proved by the next theorem, which builds on \Cref{th:convergence_shallow}.

\begin{theorem}
\label{th:convergence_deep}
Assume that $\expectt[x]{\norm{x}^2} < +\infty$ and $\sigma$ is Lipschitz continuous.
Let $(\phi_j)_{j \leq J}$ be the activation layer of an infinite-width rainbow network with distributions $(\pi_j)_{j \leq J}$ with bounded second-{} and fourth-order moments, and an output $f(x)$. Let $(\hat \phi_j)_{j \leq J}$ be the activation layers of sizes $(d_j)_{j \leq J}$ of a finite-width rainbow network approximation, with an output $\hat f(x)$. Let $k_j(x,x') = \innerr{\phi_j(x),\phi_j(x')}$ and
$\hat k_j(x,x') = \innerr{\hat\phi_j(x),\hat\phi_j(x')}$. Suppose that the sorted eigenvalues of $\expectt[x]{\phi_j(x) \, \phi_j(x)\trans}$ satisfy $\lambda_{j,m} = O(m^{-\alpha_j})$ with $\alpha_j > 1$.
Then there exists $c > 0$ which does not depend upon $(d_j)_{j\leq J}$ such that
\begin{align*}
    {\expect[W_1,\dots,W_j,x,x']{\abss{\hat k_j(x,x') - k_j(x,x')}{}^2}} &\leq c  \, \paren{\varepsilon_{j-1} + d_j\prsqrtt}^2 \\
    {\expect[W_1, \dots, W_j,x] {\normm{\hat A_j \, \hat \phi_j(x) - \phi_j(x)}^2_{H_j} }} &\leq c\, \varepsilon_j^2 \\
{\expect[W_1, \dots, W_J, x]{\abss{\hat f(x) - f (x)}{}^2}} &\leq c\,\normm{f}^2_{\hilbert_J} \, \varepsilon_J^2 ,
\end{align*}
where
\[
\varepsilon_j = \summ \ell {j}  d_\ell^{-\eta_\ell/2} ~~\mbox{with}~~\eta_\ell = \frac{\alpha_\ell - 1}{2(2\alpha_\ell - 1)} > 0.
\]
\end{theorem}

The proof is given in \Cref{app:proof-convergence-deep}. It applies iteratively \Cref{th:convergence_shallow} at each layer. As in \Cref{th:convergence_shallow}, the constant $c$ is explicit and depends polynomially on the constants involved in the hypotheses. For Gaussian weight distributions $\pi_j = {\cal N}(0,C_j)$, the theorem only requires that $\|C_j \|_\infty$ is finite for each $j \leq J$, where $\norm{\cdot}_\infty$ is the operator norm (i.e., the largest eigenvalue).

This theorem proves that at each layer, a finite-width rainbow network has an empirical kernel $\hat k_j$ which converges in mean-square to the deterministic kernel $k_j$ of the infinite-width network, when all widths $d_{\ell}$ grow to infinity. Similarly, after alignment, each activation layer $\hat \phi_j$ also converges to the activation layer $\phi_j$ of the infinite-width network. Finally, the finite-width rainbow output $\hat f$ converges to a function $f$ in the RKHS $\HH_J$ of the infinite-width network. This demonstrates that all finite-width rainbow networks implement the same deterministic function when they are wide enough. Note that any relative scaling between the layer widths is allowed, as the error decomposes as a sum over layer contributions: each layer converges independently. In particular, this includes the proportional case when the widths are defined as $d_j = s\, d_j^0$ and the scaling factor $s$ grows to infinity.

The asymptotic existence of rotations between any two trained networks has implications for the geometry of the loss landscape: if the weight distributions $\pi_j$ are unimodal, which is the case for Gaussian distributions, alignment rotations can be used to build continuous paths in parameter space between the two rainbow network weights without encountering loss barriers \citep{freeman-bruna-no-loss-barrier,draxler-no-loss-barrier,garipov-wilson-no-loss-barrier}. This could not be done with permutations, which are discrete symmetries. It proves that under the rainbow assumptions, the loss landscape of wide-enough networks has a single connected basin, as opposed to many isolated ones.

\Cref{th:convergence_deep} is a law-of-large-numbers result, which is different but complementary to the central-limit neural network Gaussian process convergence of \citet{neal-nngp,williams-nngp,nngp-sohl-dickstein,nngp-matthews}. These works state that at initialization, random finite-dimensional projections of the activations $\hat \phi_j$ converge to a random Gaussian process described by a kernel. In contrast, we show in a wider setting that the activations $\hat\phi_j$ converge to a deterministic feature vector $\phi_j$ described by a more general kernel, up to a random rotation. Note that this requires no assumptions of Gaussianity on the weights or the activations. The convergence of the kernels is similar to the results of \citet{daniely-singer-random-network-hilbert-spaces}, but here generalized to non-compositional kernels obtained with arbitrary weight distributions $\pi_j$.

\Cref{th:convergence_deep} can be considered as a multi-layer but static extension of the mean-field limit of \citet{chizat-bach-mean-field,mei-montanari-mean-field,rotskoff-vanden-eijnden-mean-field,sirignano-spiliopoulos-mean-field}. The limit is the infinite-width rainbow networks of \Cref{def:infinite_rainbow_network}. It differs from other multi-layer extensions \citep{sirignano-spiliopoulos-multilayer-mean-field,e-wojtowytsch-multilayer-mean-field,nguyen-pham-multilayer-mean-field,chen-vanden-eijnden-bruna-three-layer-mean-field,yang-feature-learning-infinite-width,bordelon-pehlevan-dynamical-mean-field} because \Cref{def:finite-rainbow-network} includes the alignment rotations $\hat A_j$. We shall not model the optimization dynamics of rainbow networks when trained with SGD, but we will make several empirical observations in \Cref{sec:validation}.

Finally, \Cref{th:convergence_deep} shows that the two assumptions of \Cref{def:finite-rainbow-network}, namely that layer dependencies are reduced to alignment rotations and that neuron weights are conditionally i.i.d.\  at each layer, imply the convergence up to rotations of network activations at each layer. We will verify numerically this convergence in \Cref{sec:validation} for several network architectures on image classification tasks, corroborating the results of \citet{raghu2017svcca} and \citet{kornblith-cka}. It does not mean that the assumptions of \Cref{def:finite-rainbow-network} are valid, and verifying them is challenging in high-dimensions beyond the Gaussian case where the weight distributions $\pi_j$ are not known. We however note that the rainbow assumptions are satisfied at initialization with $\pi_j = \normal(0, \Id)$, as \cref{eq:conditional_covariance} implies that $\hat C_j = \Id$ and thus that the weight matrices $W_j = G_j$ are independent. \Cref{th:convergence_deep} therefore applies at initialization. It is an open problem to show whether the existence of alignment rotations $\hat A_j$ is preserved during training by SGD, or whether dependencies between layer weights are indeed reduced to these rotations. Regarding (conditional) independence between neuron weights, \citet{sirignano-spiliopoulos-mean-field} show that in one-hidden-layer networks, neuron weights remain independent at non-zero but finite training times in the infinite-width limit. In contrast, a result of \citet{rotskoff-vanden-eijnden-mean-field} suggests that this is no longer true at diverging training times, as SGD leads to an approximation of the target function $f$ with a better rate than Monte-Carlo. Neuron weights at a given layer remain however (conditionally) exchangeable due to the permutation equivariance of the initialization and SGD, and therefore have the same marginal distribution. \Cref{th:convergence_deep} can be extended to dependent neuron weights $w'_{ji}$, e.g., with the more general assumption that their empirical distribution $d_j\pinv \summ i{d_j} \delta_{w'_{ji}}$ converges weakly to $\pi_j$ when the width $d_j$ increases.

\subsection{Symmetries and convolutional rainbow networks}
\label{sec:prior_convolutional_architectures}

The previous sections have defined fully-connected rainbow networks. In applications, prior information on the learning problem is often available. Practitioners then design more constrained architectures which implement inductive biases.
Convolutional networks are important examples, which enforce two fundamental properties: equivariance to translations, achieved with weight sharing, and local receptive fields, achieved with small filter supports \citep{lecun-handwritten-digit-recognition-convolution-backprop,lecun-cnn}. We first explain how equivariance to general groups may be achieved in rainbow networks. We then generalize rainbow networks to convolutional architectures.

\paragraph{Equivariant rainbow networks.}
Prior information may be available in the form of a symmetry group under which the desired output is invariant. For instance, translating an image may not change its class. We now explain how to enforce symmetry properties in rainbow networks by imposing these symmetries on the weight distributions $\pi_j$ rather than on the values of individual neuron weights $w_{ji}$. For Gaussian rainbow networks, we shall see that it is sufficient to impose that the desired symmetries commute with the weight covariances $C_j$.

Formally, let us consider $G$ a subgroup of the orthogonal group $O(d_0)$, under whose action the target function $f^\star$ is invariant: $f^\star(g x) = f^\star(x)$ for all $g \in G$. Such invariance is generally achieved progressively through the network layers. In a convolutional network, translation invariance is built up by successive pooling operations. The output $f(x)$ is invariant but intermediate activations $\phi_j(x)$ are equivariant to the group action. Equivariance is  more general than invariance. The activation map $\phi$ is equivariant if there is a representation $\rho$ of $G$ such that $\phi(gx) = \rho(g)\phi(x)$, where $\rho(g)$ is an invertible linear operator such that $\rho(gg') = \rho(g)\rho(g')$ for all $g,g' \in G$. An invariant function $f(x) = \inner{\theta, \phi(x)}$ is obtained from an equivariant activation map $\phi$ with a fixed point $\theta$ of the representation $\rho$. Indeed, if $\rho(g)\trans \theta = \theta$ for all $g \in G$, then $f(gx) = f(x)$.

We say that $\rho$ is an orthogonal representation of $G$ if $\rho(g)$ is an orthogonal operator for all $g$. When $\rho$ is orthogonal, we say that $\phi$ is orthogonally equivariant. We also say that a distribution $\pi$ is invariant under the action of $\rho$ if $\rho(g)\trans w \sim \pi$ for all $g \in G$, where $w \sim \pi$. We say that a linear operator $C$ commutes with $\rho$ if it commutes with $\rho(g)$ for all $g \in G$. Finally, a kernel $k$ is invariant to the action of $G$ if $k(gx,gx') = k(x,x')$. The following theorem proves that rainbow kernels are invariant to a group action if each weight distribution $\pi_j$ is invariant to the group representation on the activation layer $\phi_{j-1}$, which inductively defines orthogonal representations $\rho_j$ at each layer.

\begin{theorem}
\label{th:rainbow_activations_equivariance}
Let $G$ be a subgroup of the orthogonal group ${O}(d_0)$.
If all weight distribution $(\pi_j)_{j \leq J}$ are invariant to the inductively defined orthogonal representation of $G$ on their input activations, then activations $(\phi_j)_{j \leq J}$ are orthogonally equivariant to the action of $G$, and the rainbow kernels $(k_j)_{j \leq J}$ are invariant to the action of $G$. For Gaussian rainbow networks, this is equivalent to imposing that all weight covariances $(C_j)_{j \leq J}$ commute with the orthogonal representation of $G$ on their input activations.
\end{theorem}

The proof is in \Cref{app:proof_rainbow_activations_equivariance}. The result is proved by induction. If $\phi_{j}$ is orthogonally equivariant and $\pi_{j+1}$ is invariant to its representation $\rho_j$, then the next-layer activations are equivariant. Indeed, for $w \sim \pi_{j+1}$,
\begin{align*}
    \sigma\paren{\inner{w, \phi_{j}(gx) \vphantom{\rho_j(g)\trans}}}
    = \sigma\paren{\inner{w, \rho_{j}(g)\phi_{j}(x) \vphantom{\rho_j(g)\trans}}}
    = \sigma\paren{\inner{\rho_{j}(g)\trans w, \phi_{j}(x)}}
    \sim \sigma\paren{\inner{w, \phi_{j}(x) \vphantom{\rho_j(g)\trans}}},
\end{align*}
which defines an orthogonal representation $\rho_{j+1}$ on $\phi_{j+1}$. Note that any distribution $\pi_j$ which is invariant to an orthogonal representation $\rho_j$ necessarily has a covariance $C_j$ which commutes with $\rho_j$. The converse is true when $\pi_j$ is Gaussian, which shows that Gaussian rainbow networks have a maximal number of symmetries among rainbow networks with weight covariances $C_j$.

Together with \Cref{th:convergence_deep}, \Cref{th:rainbow_activations_equivariance} implies that finite-width rainbow networks can implement functions $\hat f$ which are approximately invariant, in the sense that the mean-square error $\expectt[W_1,\dots,W_J,x]{\abss{\hat f(gx) - \hat f(x)}{}^2}$ vanishes when the layer widths grow to infinity, with the same convergence rate as in \Cref{th:convergence_deep}. The activations $\hat\phi_j$ are approximately equivariant in a similar sense.
This gives a relatively easy procedure to define neural networks having predefined symmetries. The usual approach is to impose that each weight matrix $W_j$ is permutation-equivariant to the representation of the group action on each activation layer \citep{cohen-welling-group-equivariance,kondor-group-convolution}. This means that $W_j$ is a group convolution operator and hence that the rows of $W_j$ are invariant by this group action. This property requires weight-sharing or synchronization between weights of different neurons, which has been criticized as biologically implausible \citep{weight-sharing-biological-plausability-2018,weight-sharing-biological-plausability-2020,weight-sharing-biological-plausability-2021}. On the contrary, rainbow networks implement symmetries by imposing that the neuron weights are independent samples of a distribution which is invariant under the group action. The synchronization is thus only at a global, statistical level. It also provides representations with the orthogonal group, which is much richer than the permutation group, and hence increases expressivity. It comes however at the cost of an approximate equivariance for finite layer widths.

\paragraph{Convolutional rainbow networks.}
Translation-equivariance could be achieved in a fully-connected architecture by imposing stationary weight distributions $\pi_j$. For Gaussian rainbow networks, this means that weight covariances $C_j$ commute with translations, and are thus convolution operators. However, the weights then have a stationary Gaussian distribution and therefore cannot have a localized support. This localization has to be enforced with the architecture, by constraining the connectivity of the network. We generalize the rainbow construction to convolutional architectures, without necessarily imposing that the weights are Gaussian. It is achieved by a factorization of the weight layers, so that identical random features embeddings are computed for each patch of the input. As a result, all previous theoretical results carry over to the convolutional setting.

In convolutional networks, each $W_j$ is a convolution operator which enforces both translation equivariance and locality. Typical architectures impose that convolutional filters have a predefined support with an output which may be subsampled. This architecture prior can be written as a factorization of the weight matrix:
\begin{equation*}
    W_j = L_j\, P_j ,
\end{equation*}
where $P_j$ is a prior convolutional operator which only acts along space and is replicated over channels (also known as depthwise convolution), while $L_j$ is a learned pointwise (or $1 \times 1$) convolution which only acts along channels and is replicated over space. This factorization is always possible, and should not be confused with depthwise-separable convolutions \citep{sifre-rotation-scaling-scat,chollet-depthwise-separable-convs}.

Let us consider a convolutional operator $W_j$ having a spatial support of size $s_j^2$,
with $d_{j-1}$ input channels and $d_j$ output channels. The prior operator $P_j$ then extracts $d_{j-1}$ patches of size $s_j \times s_j$ at each spatial location and reshapes them as a channel vector of size $d'_{j-1} = d_{j-1} s_j^2$. $P_j$ is fixed during training and represents the architectural constraints imposed by the convolutional layer. The learned operator $L_j$ is then a $1 \times 1$ convolutional operator, applied at each spatial location across $d'_{j-1}$ input channels to compute $d_j$ output channels. This factorization reshapes the convolution kernel of $W_j$ of size $d_j \times d_{j-1} \times s_j \times s_j$ into a $1 \times 1$ convolution $L_j$ with a kernel of size $d_j \times d'_{j-1} \times 1 \times 1$. $L_j$ can then be thought as a fully-connected operator over channels that is applied at every spatial location.

The choice of the prior operator $P_j$ directly influences the learned operator $L_j$ and therefore the weight distributions $\pi_j$. $P_j$ may thus be designed to achieve certain desired properties on $\pi_j$. For instance, the operator $P_j$ may also specify predefined filters, such as wavelets in learned scattering networks \citep{separationiclr,phase-collapse}. In a learned scattering network, $P_j$ computes spatial convolutions and subsamplings, with $q$ wavelet filters having different orientations and frequency selectivity. The learned convolution $L_j$ then has $d'_{j-1} = d_{j-1} q$ input channels. This is further detailed in \Cref{app:architectures}, which explains that one can reduce the size of $L_j$ by imposing that it commutes with $P_j$, which amounts to factorizing $W_j = P_j \, L_j$ instead.

The rainbow construction of \Cref{sec:infinite_rainbow_kernel} has a straightforward extension to the convolutional case, with a few adaptations. The activations layers $\hat\phi_{j-1}$ should be replaced with $P_j \hat\phi_{j-1}$ and $W_j$ with $L_j$, where it is understood that it represents a fully-connected matrix acting along channels and replicated pointwise across space. Similarly, the weight covariances $C_j$ and its square roots $C_j\psqrtt$ are $1 \times 1$ convolutional operators which act along the channels of $P_j \hat\phi_{j-1}$, or equivalently are applied over patches of $\hat\phi_{j-1}$. Finally, the alignments $\hat A_{j-1}$ are $1 \times 1$ convolutions which therefore commute with $P_j$ as they act along different axes. One can thus still define $\hat C_j = \hat A_{j-1}\trans C_j \hat A_{j-1}$. Convolutional rainbow networks also satisfy \Cref{th:convergence_shallow,th:convergence_deep,th:rainbow_activations_equivariance} with appropriate modifications.

We note that the expression of the rainbow kernel is different for convolutional architectures. Equation (\ref{eq:rainbow_kernel_depth_j}) becomes
\[
    k_j (x,x') = \sum_u \expect[w\sim \pi_j]{ \sigma\paren{\inner{w, P_j \phi_{j-1}(x)[u]}} \sigma\parenn{\innerr{w, P_j \phi_{j-1}(x')[u]}} } ,
\]
where $P_j\phi_{j-1}(x)[u]$ is a patch of $\phi_{j-1}(x)$ centered at $u$ and whose spatial size is determined by $P_j$.
In the particular case where $\pi_j$ is Gaussian with a covariance $C_j$, the dot-product kernel in \cref{eq:dot_product_kernel_depth_j} becomes
\[
k_j (x,x') =  \sum_u  {\|z_u(x)\|\, \|z_u(x')\|}\, \kappa\paren{\frac{\lb z_u(x), z_u(x')\rb} {\|z_u(x)\|\, \|z_u(x')\|}} ~~\text{with}~~ z_u(x) = C_j^{1/2} P_j\phi_{j-1}(x)[u],
\]
The sum on the spatial location $u$ averages the local dot-product kernel values and defines a translation-invariant kernel. Observe that it differs from the fully-connected rainbow kernel (\ref{eq:dot_product_kernel_depth_j}) with weight covariances $C'_j = P_j\trans C_j P_j$, which is a global dot-product kernel with a stationary covariance. Indeed, the corresponding fully-connected rainbow networks have filters with global spatial support, while convolutional rainbow networks have localized filters. The covariance structure of depthwise convolutional filters has been investigated by \citet{trockman-kolter-covariance-convolutional-filters}.

The architecture plays an important role by modifying the kernel and hence the RKHS ${\cal H}_J$ of the output \citep{daniely-singer-random-network-hilbert-spaces}. Hierarchical convolutional kernels have been studied by \citet{mairal-ckn,anselmi-rosasco-poggio-hierarchical-kernels-groups,BiettiThese}. \citet{bietti-kernel-cnn} have proved that functions in $\hilbert_J$ are stable to the action of diffeomorphisms \citep{mallatscattering} when $P_j$ also include a local averaging before the patch extraction. When $C_j=\Id$, such kernels have been shown to efficiently approximate and learn local functions \citep{cagnetta-wyart-kernel-wide-cnn}. In that case, deep kernels with $J > 1$ hidden layers are not equivalent to shallow kernels with $J=1$ \citep{bietti-bach-dot-product-kernel-cascade}.

\section{Numerical results}
\label{sec:validation}

In this section, we validate the rainbow model on several network architectures trained on image classification tasks and make several observations on the properties of the learned weight covariances $C_j$.
As our first main result, we partially validate the rainbow model by showing that network activations converge up to rotations when the layer widths increase (\Cref{sec:convergence}). We then show in \Cref{sec:covariance_properties} that the empirical weight covariances $\hat C_j$ converge up to rotations when the layer widths increase. Furthermore, the weight covariances are typically low-rank and can be partially specified from the input activation covariances. Our second main result, in \Cref{sec:gaussianity-resampling}, is that the Gaussian rainbow model applies to scattering networks trained on the CIFAR-10 dataset. Generating new weights from the estimated covariances $C_j$ leads to similar performance than SGD training when the network width is large enough. We further show that SGD only updates the weight covariance during training while preserving the white Gaussian initialization. It suggests a possible explanation for the Gaussian rainbow model, though the Gaussian assumption seems too strong to hold for more complex learning tasks for network widths used in practice.

\subsection{Convergence of activations in the infinite-width limit}
\label{sec:convergence}

We show that trained networks with different initializations converge to the same function when their width increases. More precisely, we show the stronger property that at each layer, their activations converge after alignment to a fixed deterministic limit when the width increases. Trained networks thus share the convergence properties of rainbow networks (\Cref{th:convergence_deep}). \Cref{sec:gaussianity-resampling} will further show that scattering networks trained on CIFAR-10 indeed approximate Gaussian rainbow networks. In this case, the limit function is thus in the Gaussian rainbow RKHS (\Cref{def:infinite_rainbow_network}).

\paragraph{Architectures and tasks.}
In this paper, we consider two architectures, learned scattering networks \citep{separationiclr,phase-collapse} and ResNets \citep{resnet}, trained on two image classification datasets, CIFAR-10 \citep{cifar} and ImageNet \citep{imagenet-dataset}. 

Scattering networks have fixed spatial filters, so that their learned weights only operate across channels. This structure reduces the learning problem to channel matrices and plays a major role in the (conditional) Gaussianity of the learned weights, as we will see. The networks have $J$ hidden layers, with $J = 7$ on CIFAR-10 and $J = 10$ on ImageNet. Each layer can be written $W_j = L_j \, P_j$ where $L_j$ is a learned $1 \times 1$ convolution, and $P_j$ is a convolution with predefined complex wavelets. $P_j$ convolves each of its $d_{j-1}$ input channels with $5$ different wavelet filters ($1$ low-frequency filter and $4$ oriented high-frequency wavelets), thus generating $d'_{j-1} = 5d_{j-1}$ channels. We shall still denote $L_j$ with $W_j$ to keep the notations of \Cref{sec:infinite_rainbow_kernel}. The non-linearity $\sigma$ is a complex modulus with skip-connection, followed by a standardization (as computed by a batch-normalization). This architecture is borrowed from \citet{phase-collapse} and is further detailed in \Cref{app:architectures}. 

Our scattering network reaches an accuracy of $92\%$ on the CIFAR-10 test set. As a comparison, ResNet-20 \citep{resnet} achieves $91\%$ accuracy, while most linear classification methods based on hierarchical convolutional kernels such as the scattering transform or the neural tangent kernel reach less than $83\%$ accuracy \citep{mairal-ckn,oyallon-roto-translation-scattering,li-arora-enhanced-ntk}. On the ImageNet dataset \citep{imagenet-dataset}, learned scattering networks achieve $89\%$ top-5 accuracy \citep{separationiclr,phase-collapse}, which is also the performance of ResNet-18 with single-crop testing. 

We have made minor adjustments to the ResNet architecture for ease of analysis such as removing bias parameters (at no cost in performance), as explained in \Cref{app:architectures}. It can still be written $W_j = L_j \, P_j$ where $P_j$ is a patch extraction operator as explained in \Cref{sec:prior_convolutional_architectures}, and the non-linearity $\sigma$ is a ReLU.

\paragraph{Convergence of activations.}

\begin{figure*}[t]
  \centering
  \includegraphics[width=0.7\textwidth]{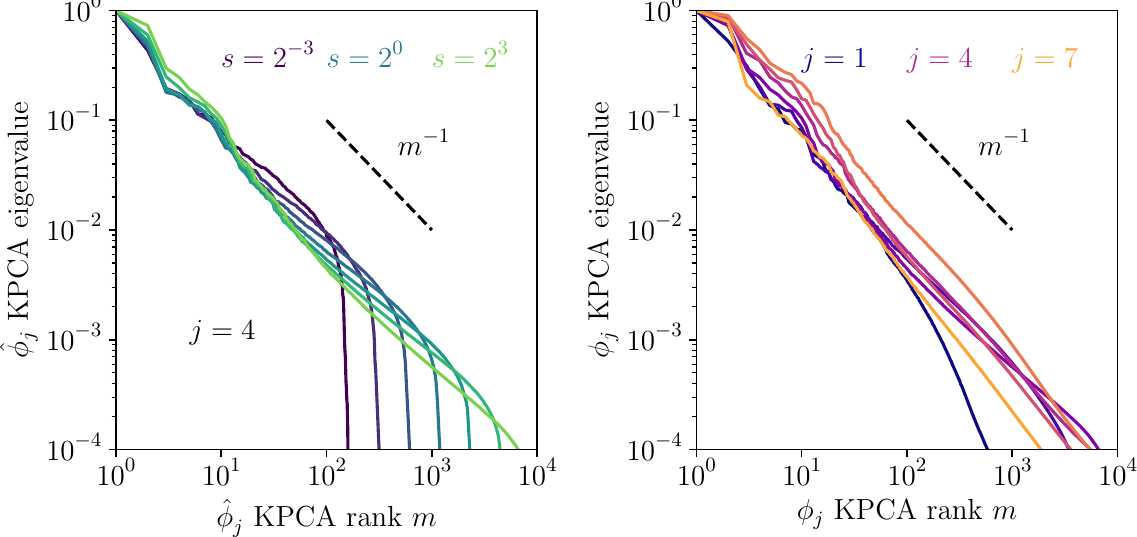}
  \caption{\small
  Convergence of spectra of activations $\hat\phi_j$ of finite-width trained scattering networks towards the feature vector $\phi_j$. The figure shows the covariance spectra of activations $\hat\phi_j$ for a given layer $j = 4$ and various width scaling $s$ (\emph{left}) and of the feature vector $\phi_j$ for the seven hidden layers $j \in \curly{1, \dots, 7}$ (\emph{right}).
  The covariance spectrum is a power law of index close to $-1$.
  }
  \label{fig:alignment_convergence}
\end{figure*}

We train several networks with a range of widths by simultaneously scaling the widths of all layers with a multiplicative factor $s$ varying over a range of $2^6 = 64$. We show that their activations $\hat\phi_j$ converge after alignment to a fixed deterministic limit $\phi_j$ when the width increases. The feature map $\phi_j$ is approximated with the activations of a large network with $s = 2^3$.

We begin illustrating the behavior of activation spectra as a function of our width-scaling parameter $s$, for seven-hidden-layer trained scattering networks on CIFAR-10. In the left panel of \Cref{fig:alignment_convergence}, we show how activation spectra vary as a function of $s$ for the layer $j=4$ which has a behavior representative of all other layers. The spectra are obtained by doing a PCA of the activations $\hat\phi_j(x)$, which corresponds to a KPCA of the input $x$ with respect to the empirical kernel $\hat k_j$. The $\hat\phi_j$ covariance spectra for networks of various widths overlap at lower KPCA ranks, suggesting well-estimated components, while the variance then decays rapidly at higher ranks. Wider networks thus estimate a larger number of principal components of the feature vector $\phi_j$. For the first layer $j=1$, this recovers the random feature KPCA results of \citet{sriperumbudur-kernel-pca-rf}, but this convergence is observed at all layers. The overall trend as a function of $s$ illustrates the infinite-width convergence. We also note that, as the width increases, the activation spectrum becomes closer to a power-law distribution with a slope of $-1$. The right panel of the figure shows that this type of decay with KPCA rank $m$ is observed at all layers of the infinite-width network $(\phi_j)_{j \leq J}$.
The power-law spectral properties of random feature activations have been studied theoretically by \citet{scetbon-harchaoui-spectrum-dot-product-kernel}, and in connection with empirical scaling laws \citep{hestness-empirical-scaling-law,kaplan-scaling-laws-llm} by \citet{spigler-geiger-wyart-scaling-laws,bordelon2020spectrum,maloney-roberts-sully-solvable-neural-scaling-laws}. Note that here we do not scale the dataset size nor training hyperparameters such as the learning rate or batch size with the network width, and a different experimental setup would likely influence the infinite-width limit \citep{yang-hyperparameter-transfer,deepmind-chinchilla}.

\begin{figure*}[t]
  \centering
  \includegraphics[width=0.75\textwidth]{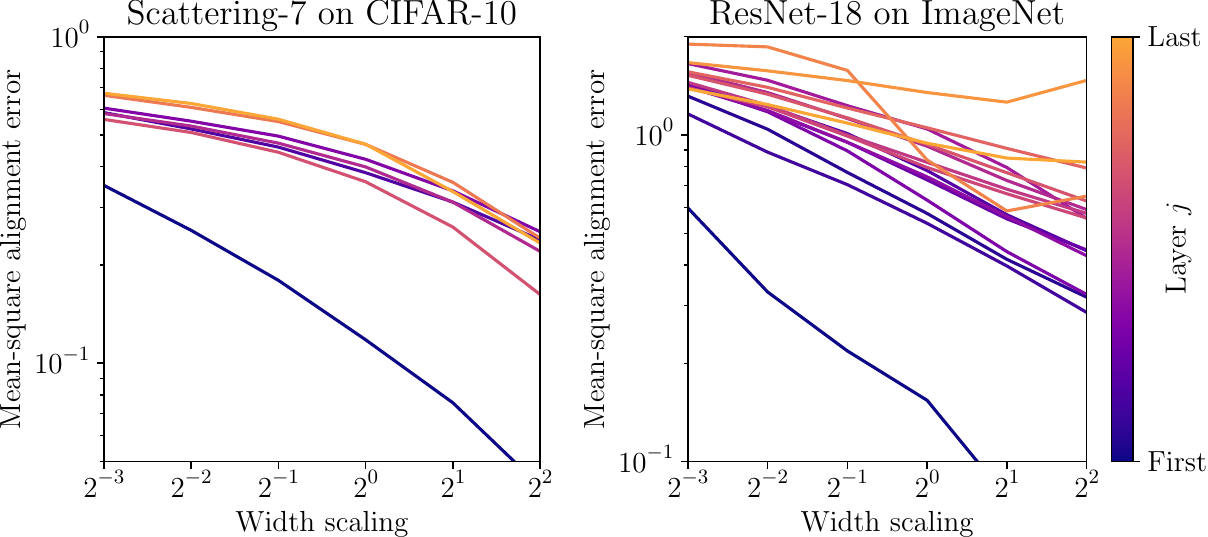}
  \caption{\small
  Convergence of activations $\hat\phi_j$ of finite-width networks towards the corresponding feature vector $\phi_j$, for scattering networks trained on CIFAR-10 (\emph{left}) and ResNet trained on ImageNet (\emph{right}). Both panels show the relative mean squared error $\expectt[x]{\normm{\hat A_j \, \hat \phi_j(x) - \phi_j(x)}{}^2} / \expectt[x]{\normm{\phi_j(x)}^2}$ between aligned activations $\hat A_j \, \hat\phi_j$ and the feature vector $\phi_j$. The error decreases as a function of the width scaling $s$ for all layers for the scattering network, and all but the last few layers for ResNet.
  }
  \label{fig:alignment_convergence_resnet}
\end{figure*}

We now directly measure the convergence of activations by evaluating the mean-square distance after alignment $\expectt[x]{\normm{\hat A_j \, \hat \phi_j(x) - \phi_j(x)}{}^2}$. The left panel of \Cref{fig:alignment_convergence_resnet} shows that it does indeed decrease when the network width increases, for all layers $j$.
Despite the theoretical convergence rate of \Cref{th:convergence_deep} vanishing when the activation spectrum exponent $\alpha_j$ approaches $1$, in practice we still observe convergence. Alignment rotations $\hat A_j$ are computed on the train set while the mean-square distance is computed on the test set, so this decrease is not a result of overfitting.
It demonstrates that scattering networks $\hat\phi_j$ approximate the same deterministic network $\phi_j$ no matter their initialization or width when it is large enough. The right panel of the figure evaluates this same convergence on a ResNet-18 trained on ImageNet. The mean-square distance after alignment decreases for most layers when the width increases. We note that the rate of decrease slows down for the last few layers. For these layers, the relative error after alignment is of the order of unity, indicating that the convergence is not observed at the largest width considered here. The overall trend however suggests that further increasing the width would reduce the error after alignment.
The observations that networks trained from different initializations have similar activations had already been made by \citet{raghu2017svcca}. \citet{kornblith-cka} showed that similarity increases with width, but with a weaker similarity measure. Rainbow networks, which we will show can approximate scattering networks, explain the source of these observations as a consequence of the law of large numbers applied to the random weight matrices with conditionally i.i.d.\ rows.

\subsection{Properties of learned weight covariances}
\label{sec:covariance_properties}

We have established the convergence (up to rotations) of the activations $\hat\phi_j$ in the infinite-width limit. Under the rainbow model, the weight matrices $W_j$ are random and thus cannot converge. However, they define estimates $\tilde C_j$ of the infinite-dimensional weight covariances $C_j$. We show that these estimates $\tilde C_j$ converge to the true covariances $C_j$ when the width increases. We then demonstrate that the covariances $C_j$ are effectively low-rank, and that their eigenspaces can be efficiently approximated by taking into account unsupervised information. The weight covariances are thus of low complexity, in the sense that they can be described with a number of parameters significantly smaller than their original size.

\paragraph{Estimation of the weight covariances.}
We estimate the weight covariances $C_j$ from the learned weights of a deep network. This network has weight matrices $W_j$ of size $d_j \times d_{j-1}$ that have been trained end-to-end by SGD. The natural empirical estimate of the weight covariance $\hat C_j$ of $W_j$ is
\begin{equation}
    \label{eq:estimation_hat_c_j}
    \hat C_j \approx d_j\pinv \, W_j\trans W_j.
\end{equation}
It computes $\hat C_j$ from $d_j$ samples, which are conditionally i.i.d.\ under the rainbow model hypothesis. Although the number $d_j$ of samples is large, their dimension $d_{j-1}$ is also large. For many architectures $d_j/d_{j-1}$ remains nearly constant and we shall consider in this section that $d_j = s \, d_j^0$, so that when the scaling factor $s$ grows to infinity $d_j/d_{j-1}$ converges to a non-zero finite limit. This creates challenges in the estimation of $\hat C_j$, as we now explain. We will see that the weight variance is amplified during training. The learned covariance can thus be modeled $\hat C_j = \Id + \hat C'_j$, where the magnitude of $\hat C'_j$ keeps increasing during training. When the training time goes to infinity, the initialization $\Id$ becomes negligible with respect to $\hat C'_j$. However, at finite training time, only the eigenvectors of $C'_j$ with sufficiently high eigenvalues have been learned consistently, and $\hat C'_j$ is thus effectively low-rank. $\hat C_j$ is then a spiked covariance matrix \citep{johnstone-largest-eigenvalue}. A large statistical literature has addressed the estimation of spiked covariances when the number of parameters $d_{j-1}$ and the number of observations $d_{j}$ increases, with a constant ratio $d_{j} / d_{j-1}$ \citep{bbp-transition,el-karoui-consistent-spectrum-spiked-covariance}. Consistent estimators of the eigenvalues of $\hat C_j$ can be computed, but not of its eigenvectors, unless we have other prior information such as sparsity of the covariance entries \citep{el-karoui-consistent-spiked-sparse-covariance} or its eigenvectors \citep{ma-spiked-covariance-sparse-eigenvectors}. In our setting, we shall see that prior information on eigenspaces of $\hat C_j$ is available from the eigenspaces of the input activation covariances. We use the empirical estimator (\ref{eq:estimation_hat_c_j}) for simplicity, but it is  not optimal. Minimax-optimal estimators are obtained by shrinking empirical eigenvalues \citep{donoho-johnstone-optimal-shrinkage-eigenvalues-spiked-covariance}.

We would like to estimate the infinite-dimensional covariances $C_j$ rather than finite-dimensional projections $\hat C_j$. Since $\hat C_j = \hat A_{j-1}\trans C_j \hat A_{j-1}$, an empirical estimate of $C_j$ is given by
\begin{equation}
    \label{eq:estimation_c_j}
   \tilde C_j = \hat A_{j-1} \hat C_j \hat A_{j-1}\trans.
\end{equation}
To compute the alignment rotation $\hat A_{j-1}$ with \cref{eq:alignment_closed_form}, we must estimate the infinite-width rainbow activations $\phi_{j-1}$. As above, we approximate $\phi_{j-1}$ with the activations $\hat\phi_{j-1}$ of a finite but sufficiently large network, relying on the activation convergence demonstrated in the previous section. We then estimate $C_j$ with \cref{eq:estimation_c_j} and $\hat C_j \approx d_j\pinv \, W_j\trans W_j$. We further reduce the estimation error of $C_j$ by training several networks of size $(d_j)_{j \leq J}$, and by averaging the empirical estimators (\ref{eq:estimation_c_j}). Note that averaging directly the estimates (\ref{eq:estimation_hat_c_j}) of $\hat C_j$ with different networks would not lead to an estimate of $C_j$, because the covariances $\hat C_j$ are represented in different bases which must be aligned. The final layer weights $\theta$ are also similarly computed with an empirical estimator from the trained weights $\hat \theta$.

\paragraph{Convergence of weight covariances.}

\begin{figure*}[t]
  \centering
  \includegraphics[width=\textwidth]{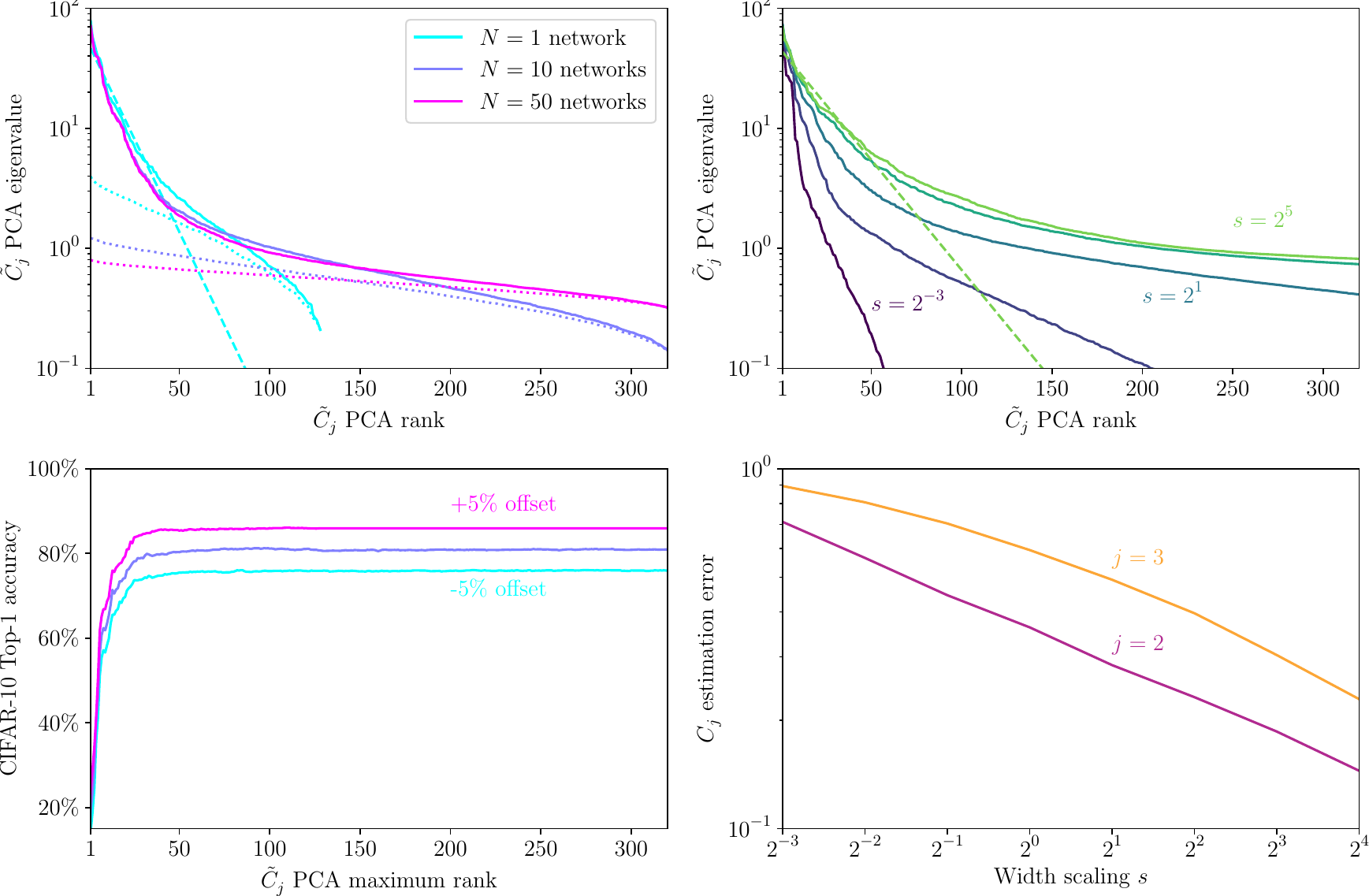}
  \caption{\small
  The weight covariance estimate $\tilde C_j$ converges towards the infinite-dimensional covariance $C_j$ for a three-hidden-layer scattering network trained on CIFAR-10. The first three panels show the behavior of the layer $j=2$.
  \emph{Upper left}: spectra of empirical weight covariances $\tilde C_j$ as a function of the network sample size $N$ showing the transition from an exponential decay (fitted by the dashed line for $N=1$) to the Marchenko-Pastur spectrum (fitted by the dotted lines).
  \emph{Lower left}: test classification performance on CIFAR-10 of the trained networks as a function of the maximum rank of its weight covariance $\tilde C_j$. Most of the performance is captured with the first eigenvectors of $\tilde C_j$. The curves for different network sample sizes $N$ when estimating $\tilde C_j$ overlap and are offset for visual purposes.
  \emph{Upper right}: spectrum of empirical weight covariances $\tilde C_j$ as a function of the network width scaling $s$. The dashed line is a fit to an exponential decay at low rank.
  \emph{Lower right}: relative distance between empirical and true covariances $\normm{\hat C_j - C_j}{}_\infty / \normm{C_j}{}_\infty$, as a function of the width scaling $s$.
  }
  \label{fig:covariance_convergence}
\end{figure*}

We now show numerically that the weight covariance estimates $\tilde C_j$ (\ref{eq:estimation_c_j}) converge to the true covariances $C_j$. This performs a partial validation of the rainbow assumptions of \Cref{def:finite-rainbow-network}, as it verifies the rotation of the second-order moments of $\pi_j$ (\ref{eq:conditional_covariance}) but not higher-moments nor independence between neurons. Due to computational limitations, we perform this verification on three-hidden-layer scattering networks trained on CIFAR-10, for which we can scale both the number of networks $N$ we can average over, and their width $s$. The main computational bottleneck here is the singular value decomposition of the cross-covariance matrix $\expectt[x]{\phi_j(x)\,\hat\phi_j(x)\trans}$ to compute the alignment $\hat A_j$, which requires $O(Ns^3)$ time and $O(Ns^2)$ memory. These shallower networks reach a test accuracy of $84\%$ at large width.

We begin by showing that empirical covariance matrices $\tilde C_j$ estimated from the weights of different networks share the same eigenspaces of large eigenvalues. To this end, we train $N$ networks of the same finite width ($s = 1$) and compare the covariances $\tilde C_j$ estimated from these $N$ networks as a function of $N$. As introduced above, the estimated covariances $\tilde C_j$ are well modeled with a spiked-covariance model.
The upper-left panel of \Cref{fig:covariance_convergence} indeed shows that the covariance spectrum interpolates between an exponential decay at low ranks (indicated by the dashed line, corresponding to the ``spikes'' resulting from training, as will be shown in \Cref{sec:gaussianity-resampling}), and a Marchenko-Pastur tail at higher ranks (indicated by dotted lines, corresponding to the initialization with identity covariance). Note that we show the eigenvalues as a function of their rank rather than a spectral density in order to reveal the exponential decay of the spike positions with rank, which was missed in previous works \citep{martin-mahoney-jmlr,thamm-staats-rosenow-rmt-weights-spectra}. The exponential regime is present even in the covariance estimated from a single network, indicating its stability across training runs, while the Marchenko-Pastur tail becomes flatter as more samples are used to estimate the empirical covariance. Here, the feature vector $\phi_j$ has been estimated with a scattering network of same width $s=1$ for simplicity of illustration.

As shown in the lower-left panel, only the exponential regime contributes to the classification accuracy of the network: the neuron weights can be projected on the first principal components of $\tilde C_j$, which correspond to the learned spikes, without harming performance. The informative component of the weights is thus much lower-dimensional ($\approx 30$) than the network width ($128$), and this dimension appears to match the characteristic scale of the exponential decay of the covariance eigenvalues. The number $N$ of trained networks used to compute $\tilde C_j$ has no appreciable effect on the approximation accuracy, which again shows that the empirical covariance matrices of all $N$ networks share this common informative component.
This presence of a low-dimensional informative weight component is in agreement with the observation that the Hessian of the loss at the end of training is dominated by a subset of its eigenvectors \citep{lecun1989optimal,hassibi1992second}. These Hessian eigenvectors could indeed be related to the weight covariance eigenvectors. Similarly, the dichotomy in weight properties highlighted by our analysis could indicate why the eigenvalue distribution of the loss Hessian separates into two distinct regimes \citep{sagun2017eigenvalues,sagun2017empirical,papyan2019measurements}: the “bulk” (with small eigenvalues corresponding to uninformative flat directions of the loss landscape) is related to the Marchenko-Pastur tail of our weight covariance spectrum and the “top” (or spiked) components correspond to the exponential regime found at the lowest ranks of the covariance spectrum.

We now demonstrate that the weight covariances $\tilde C_j$ converge to an infinite-dimensional covariance operator $C_j$ when the widths of the scattering networks increase. Here, the weight covariances $\tilde C_j$ are estimated from the weights of $N = 10$ networks with the same width scaling $s$, and we estimate $C_j$ from the weights of $N= 10$ wide scattering networks with $s = 2^5$.
We first illustrate this convergence on the spectrum of $\tilde C_j$ in the upper-right panel of \Cref{fig:alignment_convergence}. The entire spectrum of $\tilde C_j$ converges to a limiting spectrum which contains both the informative exponential part resulting from training and the uninformative Marchenko-Pastur tail coming from the initialization. The characteristic scale of the exponential regime grows with network width but converges to a finite value as the width increases to infinity. We then confirm that the estimated covariances $\tilde C_j$ indeed converge to the covariance $C_j$ when the width increases in the lower-right panel. The distance converges to zero as a power law of the width scaling. The first layer $j = 1$ has a different convergence behavior (not shown) as its input dimension does not increase with $s$.

In summary, in the context considered here, networks trained from different initializations share the same informative weight subspaces (after alignment) described by the weight covariances at each layer, and they converge to a deterministic limit when the width increases. The following paragraphs then demonstrate several properties of the weight covariances.

\paragraph{Dimensionality reduction in deep networks.}
We now consider deeper networks and show that they also learn low-rank covariances. Comparing the spectra of weights and activations reveals the alternation between dimensionality reduction with the ``colored'' weight covariances $C_j$ and high-dimensional embeddings with the ``white'' random features which are captured in the rainbow model.
We do so with two architectures: a ten-hidden-layer scattering network and a slightly modified ResNet-18 trained on ImageNet (specified in \Cref{app:architectures}), which both reach $89\%$ top-5 test accuracy.

\begin{figure*}[!t]
    \centering
    \includegraphics[width=\textwidth]{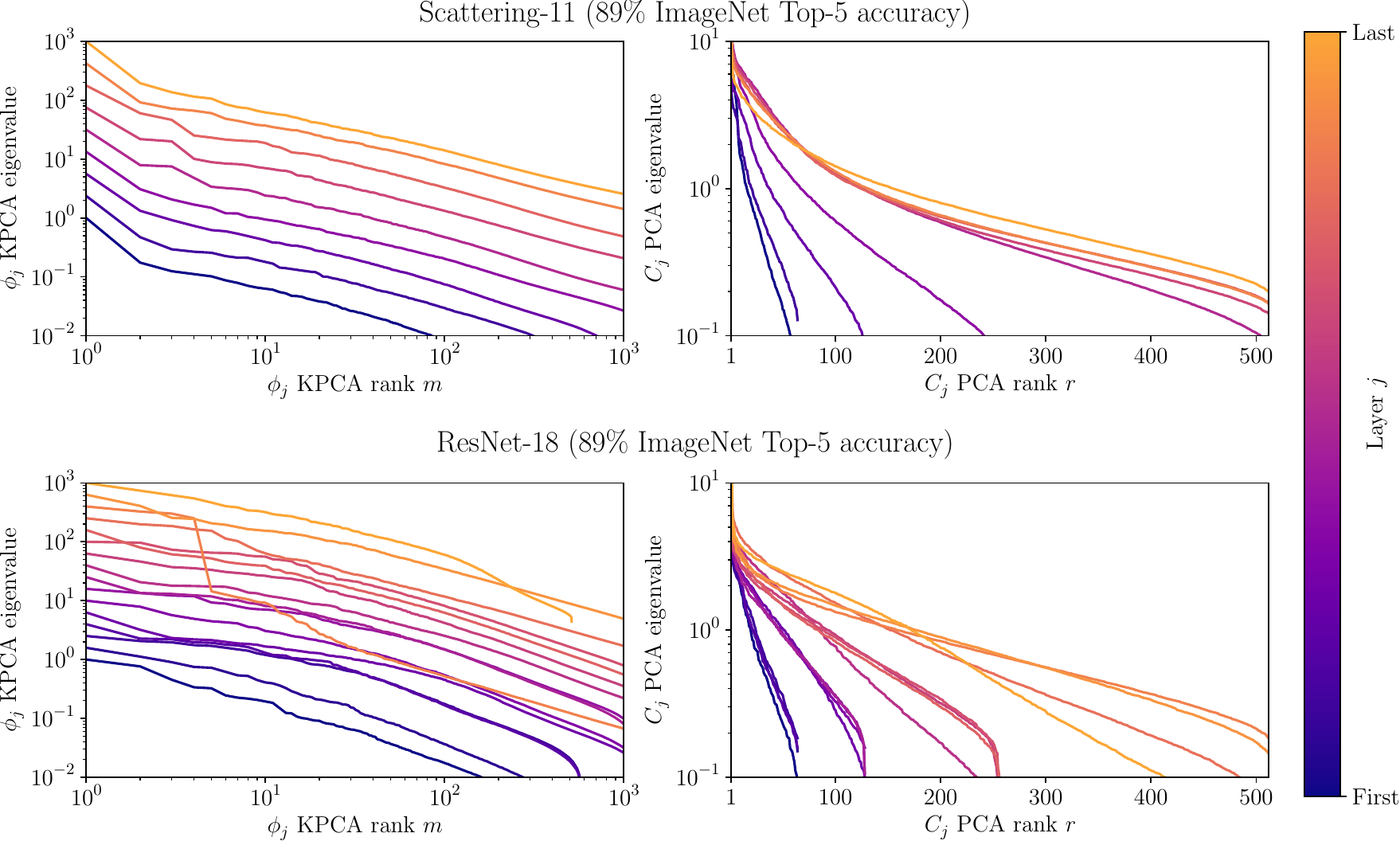}
     \caption{\small
     Covariance spectra of activations and weights of an ten-hidden-layer scattering network (\emph{top}) and ResNet-18 (\emph{bottom}) trained on ImageNet.
     In both cases, activation spectra (\emph{left}) mainly follow power-law distribution with index roughly $-1$.
     Weight spectra (\emph{right}) show a transition from an exponential decay with a characteristic scale increasing with depth to the Marchenko-Pastur spectral distribution.
     These behaviors are captured by the rainbow model.
     For visual purposes, activation and weight spectra are offset by a factor depending on $j$. In addition, we do not show the first layer nor the $1\times1$ convolutional residual branches in ResNet as they have different layer properties.
    }
    \label{fig:spectra_multilayer}
\end{figure*}

We show the spectra of covariances of activations $\phi_j$ in the left panels of \Cref{fig:spectra_multilayer} and of the weight covariances $C_j$ in the right panels. For both networks, we recover the trend that activation spectra are close to power laws of slope $-1$ and the weight spectra show a transition from a learned exponential regime to a decay consistent with the Marchenko-Pastur expectation, which is almost absent for ResNet-18. Considering them in sequence, as a function of depth, the input activations are thus high-dimensional (due to the power-law of index close to $-1$) while the subsequent weights perform a dimensionality reduction using an exponential bottleneck with a characteristic scale much smaller than the width. Next, the dimensionality is re-expanded with the non-linearity, as the activations at the next layer again have a power-law covariance spectrum. Considering the weight spectra, we observe that the effective exponential scale increases with depth, from about $10$ to $60$ for both the scattering network and the ResNet. This increase of dimensionality with depth is expected: in convolutional architectures, the weight covariances $C_j$ are only defined on small patches of activations $\phi_{j-1}$ because of the prior operator $P_j$. However, these patches correspond to a larger receptive field in the input image $x$ as the depth $j$ increases. The rank of the covariances is thus to be compared with the size of this receptive field. Deep convolutional networks thus implement a sequence of dimensionality contractions (with the learned weight covariances) and expansions (with the white random features and non-linearity). 

The successive increases and decreases in dimensionality due to the weights and non-linearity across deep network layers have been observed by \citet{recanatesi-lajoie-dimensionality-compression-expansion} with a different dimensionality measure. The observation that weight matrices of trained networks are low-rank has been made in several works which exploited it for model compression \citep{ranzato-de-freitas-predicting-low-rank-parameters,denton2014exploiting,yu2017compressing}, while the high-dimensional embedding property of random feature maps is well-known via the connection to their kernel \citep{rahimi-recht-random-features,scetbon-harchaoui-spectrum-dot-product-kernel}. The rainbow model integrates these two properties. In neuroscience, high-dimensional representations with power-law spectra have been measured in the mouse visual cortex by \citet{stringer-harris-power-law-geometry}. Such representations in deep networks have been demonstrated to lead to increased predictive power of human fMRI cortical responses \citep{elmoznino-bonner-representation-dimensionality} and generalization in self-supervised learning \citep{ghosh-richards-power-law-representation-ssl}.

\paragraph{Unsupervised approximations of weight covariances.}
The learning complexity of a rainbow network depends upon the number of parameters needed to specify the weight covariances $(C_j)_{j \leq J}$ to reach a given performance. After having shown that their informative subspace is of dimension significantly lower than the network width, we now show that this subspace can be efficiently approximated by taking into account unsupervised information (that is, information about the distribution of the input $x$ but not the target output $y$).

\begin{figure*}[!t]
  \centering
  \includegraphics[width=\textwidth]{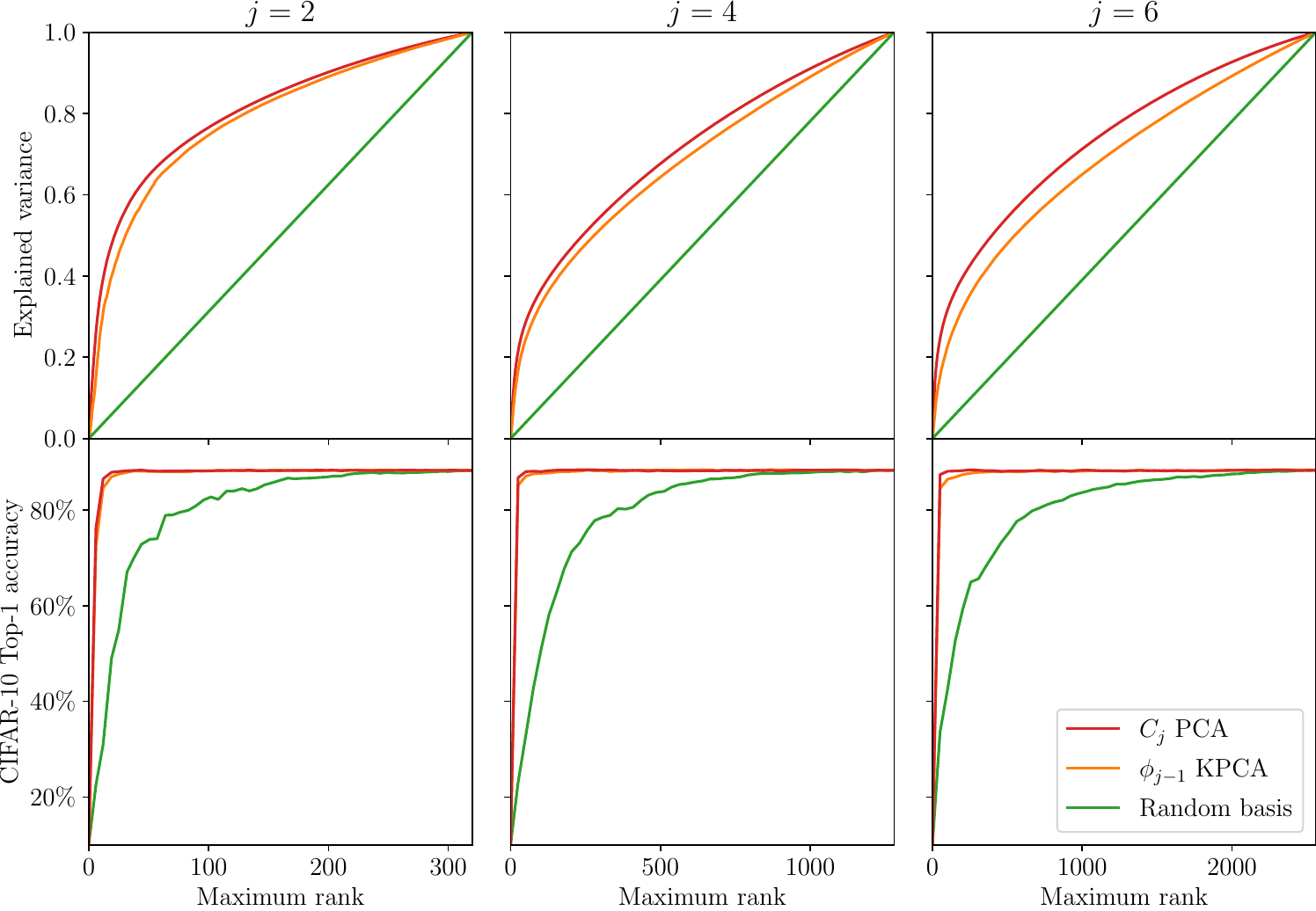}
  \caption{\small
    Unsupervised information defines low-dimensional approximations of the learned weight covariances. Each column shows a different layer $j = 2,\, 4,\, 6$ of a seven-hidden-layer scattering network trained on CIFAR-10. For each $r$, we consider projections of the network weights on the first $r$ principal components of the weight covariances (red), the kernel principal components of the input activations (orange), or random orthogonal vectors (green).
    \emph{Top}: weight variance explained by the first $r$ basis vectors as a function of $r$.
    \emph{Bottom}: classification accuracy after projection of the $j$-th layer weights on the first $r$ basis vectors, as function of $r$.
  }
  \label{fig:complexity}
\end{figure*}

We would like to define a representation of the weight covariances $C_j$ which can be accurately approximated with a limited number of parameters. We chose to represent the infinite-width activations $\phi_j$ as KPCA feature vectors, whose uncentered covariances $\expectt[x]{\phi_j(x) \, \phi_j(x)\trans}$ are diagonal. In that case, the weight covariances $C_j$ for $j > 1$ are operators defined on $H_{j-1} = \ell^2(\NN)$. It amounts to representing $C_j$ relatively to the principal components of $\phi_{j-1}$, or equivalently, the kernel principal components of $x$ with respect to $k_{j-1}$. This defines unsupervised approximations of the weight covariance $C_j$ by considering its projection on these first principal components. We now evaluate the quality of this approximation.

Here, we consider a seven-hidden-layer scattering network trained on CIFAR-10, and weight covariances estimated from $N = 50$ same-width networks. The upper panels of \Cref{fig:complexity} shows the amount of variance in $C_j$ captured by the first $m$ basis directions as a function of $m$, for three different orthogonal bases. The speed of growth of this variance as a function of $m$ defines the quality of the approximation: a faster growth indicates that the basis provides an efficient low-dimensional approximation of the covariance. The PCA basis of $C_j$ provides optimal such approximations, but it is not known before supervised training. In contrast, the KPCA basis is computed from the previous layer activations $\phi_{j-1}$ without the supervision of class label information. This corresponds to an ``unsupervised'' rainbow network which can be defined iteratively by approximating $C_1$ with the covariance of $x$, which defines a random feature representation $\phi_1(x)$, and then approximating $C_2$ with the covariance of $\phi_1(x)$, etc. \Cref{fig:complexity} demonstrates that the $\phi_{j-1}$ KPCA basis provides close to optimal approximations of $C_j$. This approximation is more effective for earlier layers, indicating that the supervised information becomes more important for the deeper layers. The lower panels of \Cref{fig:complexity} show a similar phenomenon when measuring classification accuracy instead of weight variance.

In summary, the learned weight matrices are low-rank, and a low-dimensional bottleneck can be introduced without harming performance. Further, unsupervised information (in the form of a KPCA) gives substantial prior information on this bottleneck: high-variance components of the weights are correlated with high-variance components of the activations. This observation was indirectly made by \citet{raghu2017svcca}, who showed that network activations can be projected on stable subspaces, which are in fact aligned with the high-variance kernel principal components. It demonstrates the importance of self-supervised learning within supervised learning tasks \citep{bengio-unsupervised-supervised-transfer}, and corroborates the empirical success of self-supervised pre-training for many supervised tasks. The effective number of parameters that need to be learned in a supervised manner is thus much smaller than the total number of trainable parameters. It has recently been shown that better approximations of the weight covariances can be obtained by using supervised information, in the form of the covariance of function gradients \citep{belkin-science-agop-nfa}.

\subsection{Gaussian rainbow approximations}
\label{sec:gaussianity-resampling}

We now show that the Gaussian rainbow model applies to scattering networks trained on the CIFAR-10 dataset, by exploiting the fixed wavelet spatial filters incorporated in the architecture. The Gaussian assumption thus only applies to weights along channels.
We make use of the factorization $W_j = G_j \, \hat C_j\psqrtt$ (\ref{eq:gaussian_wj_factorization}) of trained weights, where $\hat C_j$ results from an estimation of $C_j$ from several trained networks. We first show that the distribution of $G_j$ can be approximated with random matrices of i.i.d.\ normal coefficients. We then show that Gaussian rainbow networks, which replace $G_j$ with such a white Gaussian matrix, achieve similar classification accuracy as trained networks when the width is large.
Finally, we show that in the same context, the SGD training dynamics of the weight matrices $W_j$ are characterized by the evolution of the weight covariances $\hat C_j$ only, while $G_j$ remains close to its initial value. The Gaussian approximation deteriorates at small widths or on more complex datasets, suggesting that its validity regime is when the network width is large compared to the task complexity.

\paragraph{Comparison between trained weights and Gaussian matrices.}
We show that statistics of trained weights are reasonably well approximated by the Gaussian rainbow model. To do so, we train $N=50$ seven-hidden-layer scattering networks and estimate weight covariances $(C_j)_{j \leq J}$ by averaging \cref{eq:estimation_c_j} over the trained networks as explained in \Cref{sec:covariance_properties}. We then retrieve $G_j = W_j \, \hat C_j\prsqrtt$ with $\hat C_j = \hat A_j\trans C_j \hat A_j$ as in \cref{eq:conditional_covariance}. Note that we use a single covariance $C_j$ to whiten the weights of all $N$ networks: this will confirm that the covariances of weights of different networks are indeed related through rotations, as was shown in \Cref{sec:covariance_properties} through the convergence of weight covariance estimates. The rainbow feature vectors $(\phi_j)_{j \leq J}$ at each layer are approximated with the activations of one of the $N$ networks.

\begin{figure*}[t]
  \centering
  \includegraphics[width=\textwidth]{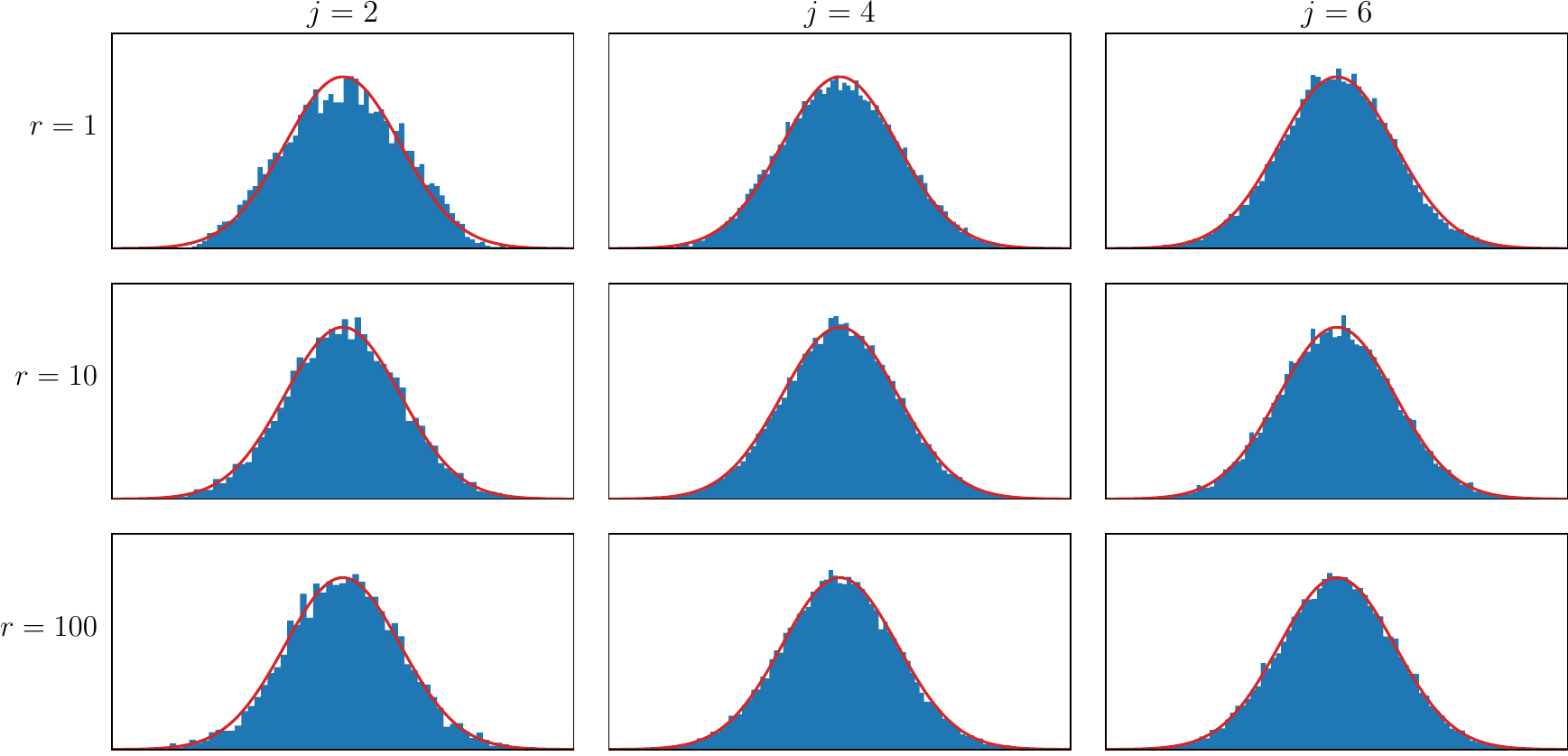}
  \caption{\small
    Marginal distributions of the weights of $N=50$ seven-hidden-layer scattering networks trained on CIFAR-10. The weights at the $j$-th layer $(w_{ji})_{i \leq d_j}$ of the $N$ networks are projected along the $r$-th eigenvector of $C_j$ and normalized by the square root of the corresponding eigenvalue. The distribution of the $Nd_j$ projections (blue histograms) is approximately normal (red curves). Each column shows a different layer $j$, and each row shows a different rank $r$.
  }
  \label{fig:gaussianity_histograms}
\end{figure*}

As a first (partial) Gaussianity test, we compare marginal distributions of whitened weights $G_{j}$ with the expected normal distribution in \Cref{fig:gaussianity_histograms}. We present results for a series of layers ($j=2,4,6$) across the network. Other layers present similar results, except for $j=1$ which has more significant deviations from Gaussianity (not shown), as its input dimension is constrained by the data dimension. We shall however not focus on this first layer as we will see that it can still be replaced by Gaussian realizations when generating new weights. The weights at the $j$-th layer $(w_{ji})_{i \leq d_j}$ of the $N$ networks are projected along the $r$-th eigenvector of $C_j$ and normalized by the square root of the corresponding eigenvalue. This global view shows that specific one-dimensional marginals are reasonably well approximated by a normal distribution.
We purposefully remain not quantitative, as the goal is not to demonstrate that trained weights are statistically indistinguishable from Gaussian realizations (which is false), but to argue that the latter is an acceptable model for the former.

\begin{figure*}[t]
  \centering
  \includegraphics[width=\textwidth]{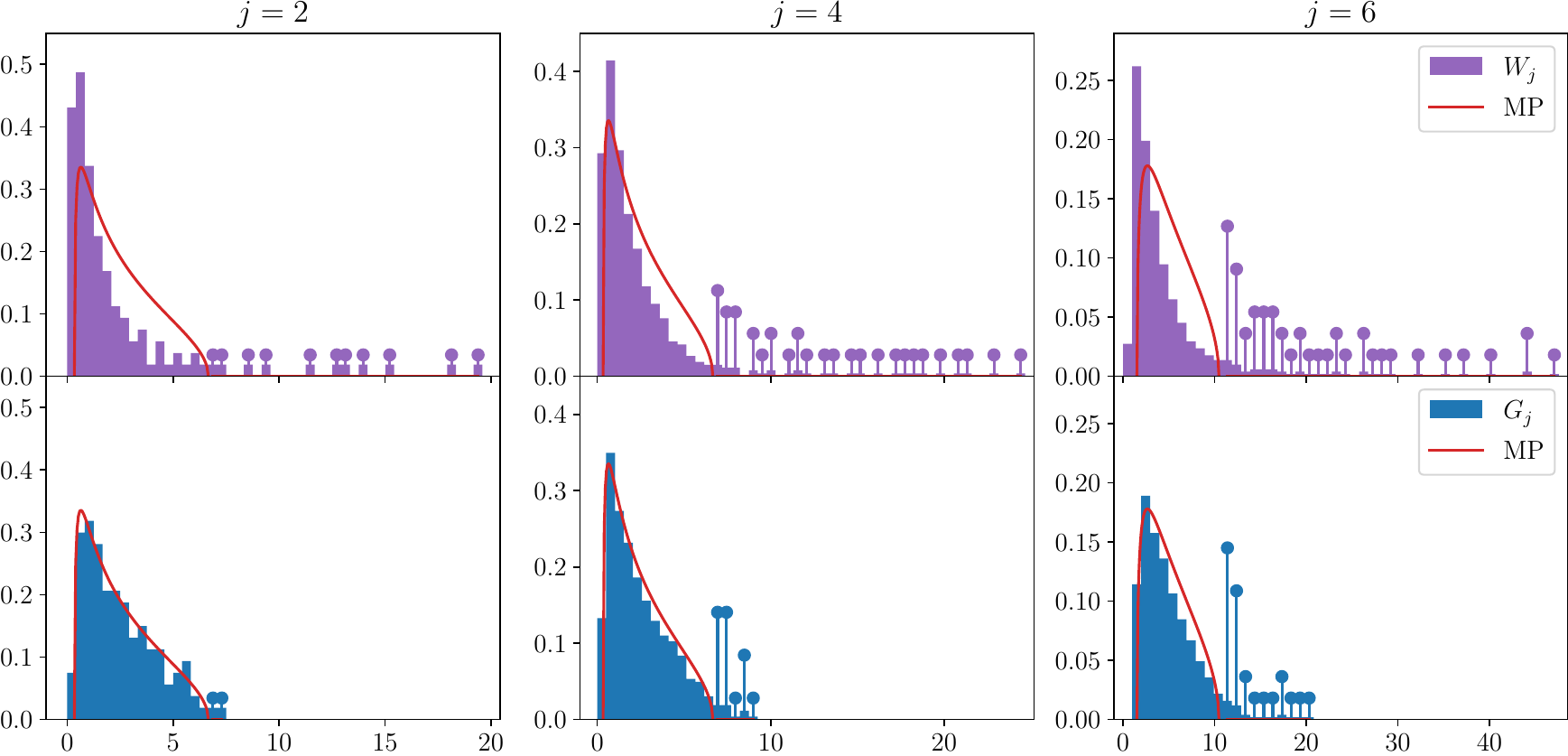}
  \caption{\small
  Spectral density of empirical covariances of trained (\emph{top}) and whitened weights (\emph{bottom}). Eigenvalues outside the support of the Marchenko-Pastur distribution (shown in red) are indicated with spikes of amplitude proportional to their bin count. After whitening, the number of outliers are respectively $2\%$, $4\%$, and $8\%$ for the layers $j=2$, $4$, and $6$.
  }
  \label{fig:gaussianity_mp}
\end{figure*}

To go beyond one-dimensional marginals, we now compare in the bottom panels of \Cref{fig:gaussianity_mp} the spectral density of the whitened weights $G_j$ to the theoretical Marchenko-Pastur distribution \citep{marchenko-pastur}, which describes the limiting spectral density of matrices with i.i.d.\ normal entries. We note a good agreement for the earlier layers, which deteriorates for deeper layers (as well as the first layer, not shown, which again has a different behavior). Importantly, the proportion of eigenvalues outside the Marchenko-Pastur support is arguably negligible ($< 10\%$ at all layers), which is not the case for the non-whitened weights $W_j$ (upper panels) where it can be $> 25\%$ for $j = 6$. As observed by \citet{martin-mahoney-jmlr} and \citet{thamm-staats-rosenow-rmt-weights-spectra}, trained weights have non-Marchenko-Pastur spectral statistics. Our results show that these deviations are primarily attributable to correlations introduced by the non-identity covariance matrices $C_j$, as opposed to power-law distributions as hypothesized by \citet{martin-mahoney-jmlr}. We however note that due to the universality of the Marchenko-Pastur distribution, even a perfect agreement is not sufficient to claim that trained networks have conditionally Gaussian weights. It merely implies that the Gaussian rainbow model provides a satisfactory description of a number of weight statistical properties. Despite the observed deviations from Gaussianity at later layers, we now show that generating new Gaussian weights at all layers simultaneously preserves most of the classification accuracy of the network.

\paragraph{Performance of Gaussian rainbow networks.}

\begin{figure*}[t]
  \centering
  \includegraphics[width=.5\textwidth]{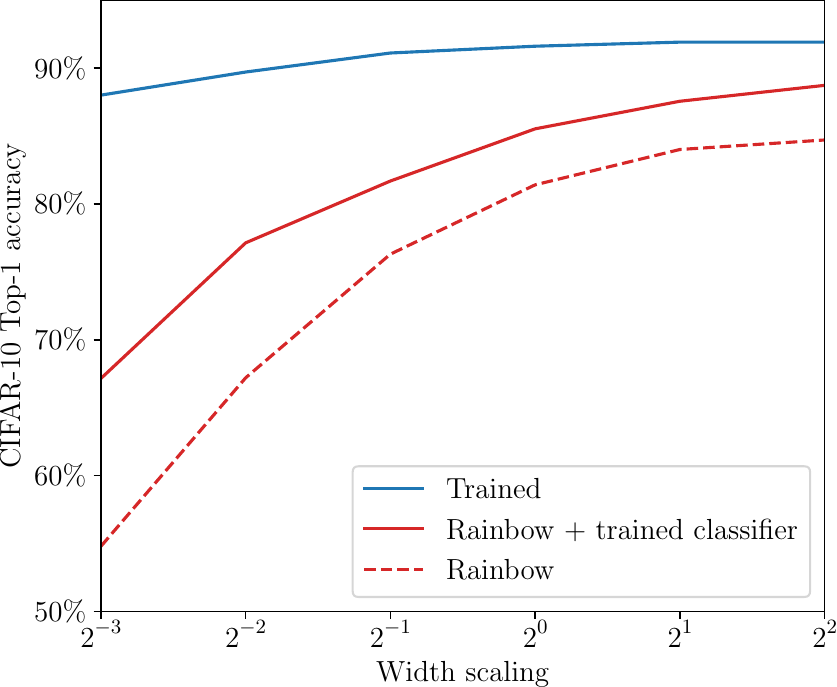}
   \caption{\small Performance of seven-hidden-layer scattering networks on CIFAR-10 as a function of network width
  for a trained network (blue), its rainbow network approximation with and without classifier retraining (red solid and dashed). The larger the width, the better the sampled rainbow model approximates the original network.
}
  \label{fig:resampling}
\end{figure*}

While the above tests indicate some level of validation that the whitened weights $G_{j}$ are matrices with approximately i.i.d.\ normal entries, it is not statistically feasible to demonstrate that this property is fully satisfied in high-dimensions. We thus sample network weights from the Gaussian rainbow model and verify that most of the performance can be recovered. This is done with the procedure described in \Cref{def:finite-rainbow-network}, using the covariances $C_j$, rainbow activations $\phi_j$ and final layer weights $\theta$ here estimated from a single trained network (having shown in \Cref{sec:convergence,sec:covariance_properties} that all networks define similar rainbow parameters if they are wide enough). New weights $W_j$ are sampled iteratively starting from the first layer with a covariance $\hat C_j = \hat A_{j-1}\trans C_j \hat A_{j-1}$, after computing the alignment rotation $\hat A_{j-1}$ between the activations $\hat\phi_{j-1}(x)$ of the partially sampled network and the activations $\phi_{j-1}(x)$ of the trained network. The alignment rotations are computed using the CIFAR-10 train set, while network accuracy is evaluated on the test set, so that the measured performance is not a result of overfitting.

We perform this test using a series of seven-hidden-layer scattering networks trained on CIFAR-10 with various width scalings. We present results in \Cref{fig:resampling} for two sets of Gaussian rainbow networks: a first set for which both the convolutional layers and the final layer are sampled from the rainbow model (which corresponds to aligning the classifier of the trained model to the sampled activations $\hat \phi_J(x)$), and another set for which we retrain the classifier after sampling the convolutional layers (which preserves the Gaussian rainbow RKHS). We observe that the larger the network, the better it can be approximated by a Gaussian rainbow model. At the largest width considered here, the Gaussian rainbow network achieves $85\%$ accuracy and $89\%$ with a retrained classifier, and recovers most of the performance of the trained network which reaches $92\%$ accuracy. This performance is non-trivial, as it is beyond most methods based on non-learned hierarchical convolutional kernels which obtain less than $83\%$ accuracy \citep{mairal-ckn,oyallon-roto-translation-scattering,li-arora-enhanced-ntk}. This demonstrates the importance of the learned weight covariances $C_j$, as has been observed by \citet{pandey-harris-colored-random-features-sensory-encoding} for modeling sensory neuron receptive fields. It also demonstrates that the covariances $C_j$ are sufficiently well-estimated from a single network to preserve classification accuracy. We note however that \citet{shankar-recht-neural-kernels-performance} achieve a classification accuracy of $90\%$ with a non-trained kernel corresponding to an infinite-width convolutional network, and \citet{thiry-patches} with a data-dependent convolutional kernel.

A consequence of our results is that these trained scattering networks have rotation invariant non-linearities, in the sense that the non-linearity can be applied in random directions, provided that the next layer is properly aligned. This comes in contrast to the idea that neuron weights individually converge to salient features of the input data. For large enough networks, the relevant information learned at the end of training is therefore not carried by individual neurons but encoded through the weight covariances $C_j$.

For smaller networks, the covariance-encoding property no longer holds, as \Cref{fig:resampling} suggests that trained weights becomes non-Gaussian. Networks trained on more complex tasks might require larger widths for the Gaussian rainbow approximation to be valid. We have repeated the analysis on scattering networks trained on the ImageNet dataset \citep{imagenet-dataset}, which reveals that the Gaussian rainbow approximation considered here is inadequate at widths used in practice. This is corroborated by many empirical observations of (occasional) semantic specialization in deep networks trained on ImageNet \citep{olah-distill-feature-viz,bau-understand-individual-units-in-deep-networks,dobs-kanwisher-specialization-in-deep-networks}. A promising direction is to consider Gaussian mixture rainbow models, as used by \citet{dubreuil-valente-ostojic-gaussian-mixture-resampling} to model the weights of linear RNNs. Finally, we note that the Gaussian approximation also critically rely on the fixed wavelet spatial filters of scattering networks. Indeed, the spatial filters learned by standard CNNs display frequency and orientation selectivity \citep{alexnet} which cannot be achieved with a single Gaussian distribution, and thus require adapted weight distributions $\pi_j$ to be captured in a rainbow model.

\paragraph{Training dynamics.} The rainbow model is a static model, which does not characterize the evolution of weights from their initialization during training. We now describe the SGD training dynamics of the seven-hidden-layer scattering network trained on CIFAR-10 considered above. This dynamic picture provides an empirical explanation for the validity of the Gaussian rainbow approximation.

We focus on the $j$-th layer weight matrix $W_{j}(t)$ as the training time $t$ evolves. To measure its evolution, we consider its projection along the principal components of the final learned covariance $\hat C_j$. More precisely, we project the $d_j$ neuron weights $w_{ji}(t)$, which are the rows of $W_j(t)$, in the direction of the $r$-th principal axis $e_{jr}$ of $\hat C_j$. This gives a vector $u_r(t) \in \RR^{d_j}$ for each PCA rank $r$ and training time $t$, dropping the index $j$ for simplicity:
\begin{align*}
  u_r(t) = \paren{\innerr{w_{ji}(t), e_{jr}}}_{i \leq d_j}.
\end{align*}
Its squared magnitude is proportional to the variance of the neuron weights along the $r$-th principal direction, which should be of the order of $1$ at $t=0$ due to the white noise initialization, and evolves during training to reach the corresponding $\hat C_j$ eigenvalue. On the opposite, the direction of $u_r(t)$ encodes the sampling of the marginal distribution of the neurons along the $r$-th principal direction: a large entry $u_r(t)[i]$ indicates that neuron $i$ is significantly correlated with the $r$-th principal component of $\hat C_j$. This view allows considering the evolution of the weights $W_j(t)$ separately for each principal component $r$. It offers a simpler view than focusing on each individual neuron $i$, because it gives an account of the population dynamics across neurons. It separates the weight matrix by columns $r$ (in the weight PCA basis) rather than rows $i$. We emphasize that we consider the PCA basis of the final covariance $\hat C_j$, so that we analyze the training dynamics along the fixed principal axes $e_{jr}$ which do not depend on the training time $t$.

\begin{figure*}[t]
  \centering
  \includegraphics[width=\textwidth]{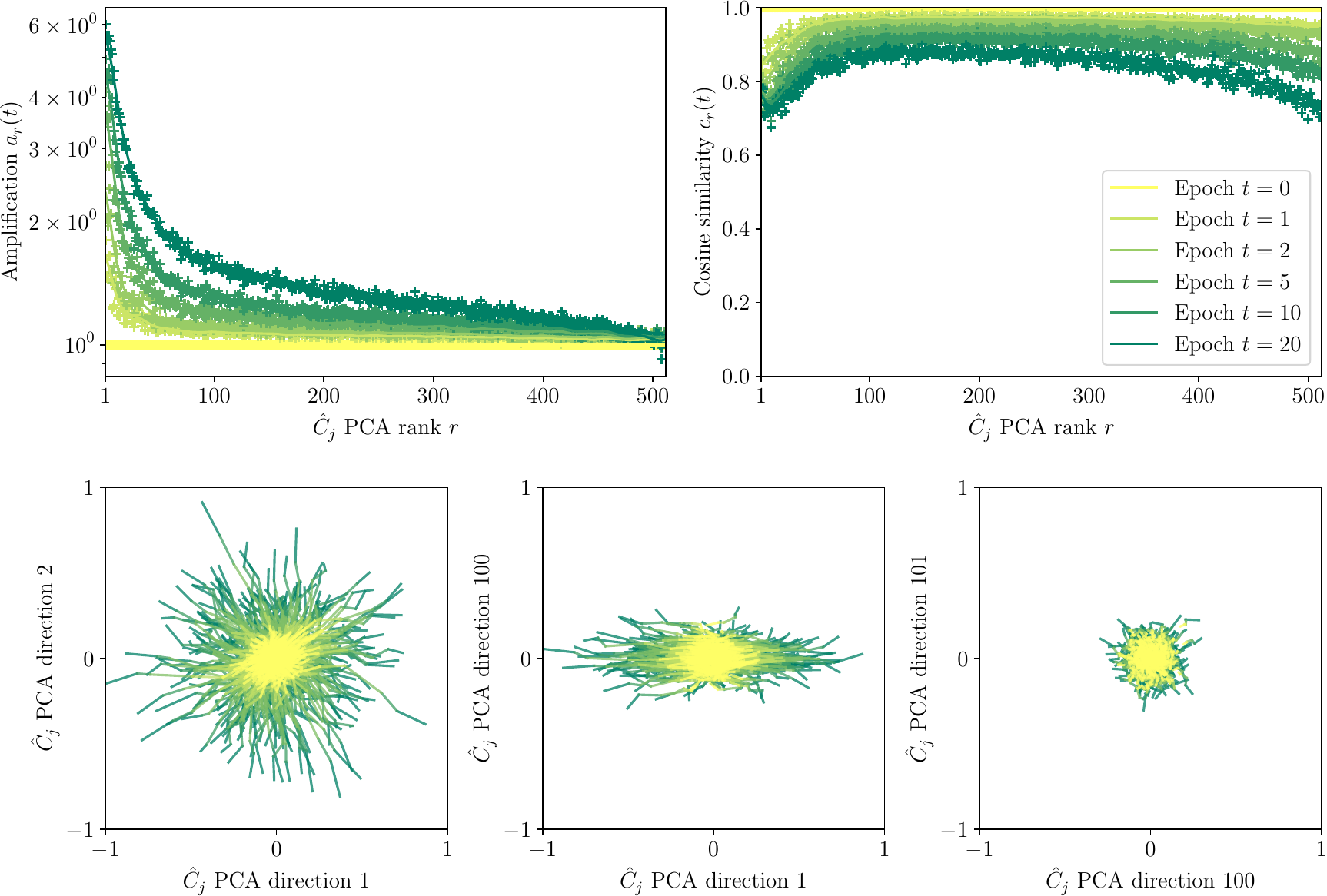}
  \caption{
  \small
  The learning dynamic of a seven-hidden-layer scattering network trained on CIFAR-10 is mainly a low-dimensional linear amplification effect that preserves most of the positional information of the initialization.
  We present results for layer $j=4$  (similar behavior is observed for the other layers).
  \emph{Upper left:} amplification (overall stretch) of the weight variance as a function of rank.
  \emph{Upper right:} cosine similarity (internal motion) as a function of rank.
  \emph{Lower panels:} projections of individual neurons along pairs of principal components. Each neuron is represented as a point in the plane, whose trajectory during training is shown as a connected line
  (color indicates training time).
  }
  \label{fig:sgd_dynamics}
\end{figure*}

We now characterize the evolution of $u_r(t)$ during training for each rank $r$. We separate changes in magnitude, which correspond to changes in weight variance (overall stretch), from changes in direction, which correspond to internal motions of the neurons which preserve their variance. We thus define two quantities to compare $u_r(t)$ to its initialization $u_r(0)$, namely the amplification ratio $a_r(t)$ and cosine similarity $c_r(t)$:
\begin{equation}
a_r(t) = \frac{\normm{u_r(t)}}{\normm{u_r(0)}}~~\text{and}~~
c_r(t) = \frac{\innerr{ u_r(t), u_r(0)} }{\normm{u_r(t)}\, \normm{u_r(0)}} .
\end{equation}
We evaluate these quantities using our seven-hidden-layer scattering network trained on CIFAR-10. In \Cref{fig:sgd_dynamics}, we present the results for the intermediate layer $j=4$ (similar behavior is observed for the other layers). We show the two quantities $a_r(t)$ and $c_r(t)$ in the top row of \Cref{fig:sgd_dynamics} as a function of the training epoch $t$. We observe that the motion of the weight vector is mainly an amplification effect operating in a sequence starting with the first eigenvectors, as the cosine similarity remains of order unity. Given the considered dimensionality ($d_j=512$), the observed departure from unity is rather small: the solid angle subtended by this angular change of direction covers a vanishingly small surface of the unit sphere in $d_j$ dimensions. We thus have $u_r(t) \approx a_r(t)\, u_r(0)$.

These results show that the weight evolution can be written
\begin{align*}
  W_j(t) \approx G_j \, \hat C_j\psqrtt(t),
\end{align*}
where $G_j = W_j(0)$ is the initialization and the weight covariance $\hat C_j(t)$ evolves by amplification in its fixed PCA basis:
\begin{equation*}
  \hat C_j(t) = \sum_r a_r(t)^2 \, e_{jr} e_{jr}\trans.
\end{equation*}
In other words, the weight evolution during training is an ensemble motion of the neuron population, with negligible internal motion of individual neurons relative to the population: training amounts to learning the weight covariance. Surprisingly, the weight configuration at the end of training thus retains most of the information of its random initialization: the initial configuration can be practically recovered by whitening the trained weights. In addition, the stochasticity introduced by SGD and data augmentation appears to be negligible, as it does not affect the relative positions of individual neurons during training. This observation has two implications. First, the alignment rotations $\hat A_j$ which describe the trained network relative to its infinite-width rainbow counterpart (as $\hat\phi_j \approx \hat A_j\trans \phi_j$) are entirely determined by the initialization. Second, it provides an empirical explanation for the validity of the Gaussian rainbow approximation. While this argument seems to imply that the learned weight distributions $\pi_j$ depend significantly on the initialization scheme, note that significantly non-Gaussian initializations might not be preserved by SGD or could lead to poor performance.

The bottom row of \Cref{fig:sgd_dynamics} illustrates more directly the evolution of individual neurons during training. Although each neuron of $W_j(t)$ is described by a $d_{j-1}$-dimensional weight vector, it can be projected along two principal directions to obtain a two-dimensional picture. We then visualize the trajectories of each neuron projected in this plane. The trajectories are almost straight lines, as the learning dynamics only amplify variance along the principal directions while preserving the relative positions of the neurons. Projections on principal components of higher ranks give a more static picture as the amplification along these directions is smaller.

A large literature has characterized properties of SGD training dynamics. Several works have observed that dynamics are linearized after a few epochs \citep{fort-cho-geras-break-even-point-hessian-loss,leclerc-mardy-two-regimes-of-training}, so that the weights remain in the same linearly connected basin thereafter \citep{frankle-carbin-child-spawning-linear-barrier}. It has also been shown that the empirical neural tangent kernel evolves mostly during this short initial phase \citep{fort-ganguli-ntk-rapid-evolution} and aligns itself with discriminative directions \citep{baratin-lajoie-vincent-lacoste-julien-ntk-alignment-amplification,atanasov-bordelon-pehlevan-kernel-alignment-amplification}. Our results indicate that this change in the neural tangent kernel is due to the large amplification of the neuron weights along the principal axes of $\hat C_j$, which happen early during training.
The observation that neural network weights have a low-rank departure from initialization has been made in connection with eigenvectors of the loss Hessian by \citet{gur-ari-sgd-tiny-subspace}, in the lazy regime by \citet{thamm-staats-rosenow-rmt-weights-spectra}, for linear RNNs by \citet{schuessler-dubreuil-ostojic-low-rank-rnn}, and for large language-model adaptation by \citet{hu-allen-zhu-low-rank-adaptation}. The sequential emergence of the weight principal components has been derived theoretically in linear networks by \citet{saxe-clelland-ganguli-linear-network-dynamics,saxe-clelland-ganguli-semantic-linear-network}.

\section{Conclusion}
\label{sec:discussion}

We have introduced rainbow networks as a model of the probability distribution of weights of trained deep networks. The rainbow model relies on two assumptions. First, layer dependencies are reduced to alignment rotations. Second, neurons are independent when conditioned on the previous layer weights. Under these assumptions, trained networks converge to a deterministic function in the corresponding rainbow RKHS when the layer widths increase. We have verified numerically the convergence of activations after alignment for scattering networks and ResNets trained on CIFAR-10 and ImageNet. We conjecture that this convergence conversely implies the rotation dependency assumption of the rainbow model. We have verified this rotation on the second-order moments of the weights through the convergence of their covariance after alignment (for scattering networks trained on CIFAR-10 due to computational limitations).

The data-dependent kernels which describe the infinite-width rainbow networks, and thus their functional properties, are determined by the learned distributions $\pi_j$. Mathematically, we have shown how the symmetry properties of these distributions are transferred on the network. Numerically, we have shown that their covariances $C_j$ compute projections in a low-dimensional ``informative'' subspace that is shared among networks, is low-dimensional, and can be approximated efficiently with an unsupervised KPCA. It reveals that networks balance low learning complexity with high expressivity by computing a sequence of reductions and increases in dimensionality.

In the Gaussian case, the distributions $\pi_j$ are determined by their covariances $C_j$. We have validated that factorizing the learned weights with fixed wavelet filters is sufficient to obtain Gaussian rainbow networks on CIFAR-10, using scattering networks. In this setting, we can generate new weights and have shown that the weight covariances $C_j$ are sufficient to capture most of the performance of the trained networks. Further, the training dynamics are reduced to learning these covariances while preserving memory of the initialization in the individual neuron weights.

Our work has several limitations. First, we have not verified the rainbow assumptions of rotation dependence between layers beyond second-order moments, and conditional independence between neurons beyond the Gaussian case. A complete model would incorporate the training dynamics and show that such statistical properties are satisfied at all times. Second, our numerical experiments have shown that the Gaussian rainbow approximation of scattering networks gradually degrades when the network width is reduced. 
When this approximation becomes less accurate, it raises the question whether incorporating more prior information in the architecture could lead to Gaussian rainbow networks. 
Finally, even in the Gaussian case, the rainbow model is not completely specified as it requires to estimate the weight covariances $C_j$ from trained weights. A major mathematical issue is to understand the properties of the resulting rainbow RKHS which result from properties of these weight covariances.

By introducing the rainbow model, this work provides new insights towards understanding the inner workings of deep networks.

\acks{This work was partially supported by a grant from the PRAIRIE 3IA Institute of the French ANR-19-P3IA-0001 program. BM acknowledges support from the David and Lucile Packard Foundation. We thank the Scientific Computing Core at the Flatiron Institute for the use of their computing resources. BM thanks Chris Olah and Eric Vanden-Eijnden for inspiring discussions. FG would like to thank Francis Bach and Gabriel Peyré for helpful pointers for the proof of \Cref{th:convergence_shallow}. We also thank Kameron Decker Harris, Nathanaël Cuvelle-Magar, and Etienne Lempereur for feedback on the manuscript.}

\appendix

\section{Proof of \protect\Cref{th:convergence_shallow}}
\label{app:proof-convergence-shallow}

\newcommand\Ltwo{L^2(\mu)}
\newcommand\bw{\mathrm{BW}}

We prove a slightly more general version of \Cref{th:convergence_shallow} which we will need in the proof of \Cref{th:convergence_deep}. We allow the input $x$ to be in a possibly infinite-dimensional separable Hilbert space $H_0$ (the finite-dimensional case is recovered with $H_0 = \RR^{d_0})$. We shall assume that the random feature distribution $\pi$ has bounded second-{} and fourth-order moments in the sense of \Cref{sec:infinite_rainbow_kernel}: it admits a bounded uncentered covariance operator $C = \expectt[w\sim\pi]{ww\trans}$ and $\expectt[w\sim\pi]{\parenn{w\trans T w}^2} < +\infty$ for every trace-class operator $T$ on $H_0$. Without loss of generality, we assume that the non-linearity $\sigma$ is $1$-Lipschitz and that $\sigma(0) = 0$. These last assumptions simplify the constants involved in the analysis. They can be satisfied for any $L$-Lipschitz non-linearity $\sigma$ by replacing it with $(\sigma - \sigma(0))/L$, which does not change the linear expressivity of the network.

We give the proof outline in \Cref{sec:convergence-proof-outline}. It relies on several lemmas, which are proven in \Cref{sec:kernel-integral-operators,sec:proof-bures-wasserstein,sec:proof-bures-wasserstein-regularization,sec:proof-regularization-term-computation}. We write $\norm{\cdot}_\infty$ the operator norm, $\norm{\cdot}_2$ the Hilbert-Schmidt norm, and $\norm{\cdot}_1$ the nuclear (or trace) norm.

\subsection{Proof outline}
\label{sec:convergence-proof-outline}

The convergence of the activations $\hat\varphi(x)$ to the feature vector $\varphi(x)$ relies on the convergence of the empirical kernel $\hat k$ to the asymptotic kernel $k$. We thus begin by reformulating the mean-square error $\expectt[x]{\normm{\hat A\, \hat \varphi(x) - \varphi(x)}{}_{H}^2}$ in terms of the kernels $\hat k$ and $k$. More precisely, we will consider the integral operators $\hat T$ and $T$ associated to the kernels. These integral operators are the infinite-dimensional equivalent of Gram matrices $(k(x_i, x_{i'}))_{1 \leq i,i' \leq n}$.

Let $\mu$ be the distribution of $x$. We define the integral operator $T \colon \Ltwo \to \Ltwo$ associated to the asymptotic kernel $k$ as
\begin{align*}
    \paren{T f}(x) = \expect[x']{k(x,x') \, f(x')},
\end{align*}
where $x'$ is an i.i.d.\ copy of $x$ and $\mu$ is the law of $x$. Similarly, we denote $\hat T$ the integral operator defined by $\hat k$. Their standard properties are detailed in the next lemma. Moreover, the definition of $\hat T$ entails that it is the average of $d_1$ i.i.d.\ integral operators defined by the individual random features $(w_i)_{i \leq d_1}$ of $\hat\varphi$. The law of large numbers then implies a mean-square convergence of $\hat T$ to $T$, as proven in the following lemma.

\begin{lemma}
    \label{th:kernel_mean_square_convergence}
    $T$ and $\hat T$ are trace-class non-negative self-adjoint operators on $\Ltwo$, with
    \begin{equation*}
        \tr(T) \leq \norm{C}_\infty \, \expectt[x]{\norm{x}^2}.    
    \end{equation*}
    The eigenvalues of $T$ and $\hat T$ are the same as their respective activation covariance matrices $\expectt[x]{\varphi(x)\,\varphi(x)\trans}$ and $\expectt[x]{\hat\varphi(x) \, \hat\varphi(x)\trans}$. Besides, it holds that $\expectt[W]{\hat T} = T$ and
    \begin{equation*}
        \sqrt{\expect[W]{\normm{\hat T - T}{}_2^2}} = \sqrt{\expect[W,x,x']{\abss{\hat k(x,x') - k(x,x')}{}^2}} = c \, d_1\prsqrtt,
    \end{equation*}
    with some constant $c < +\infty$.
\end{lemma}
Note that the last statement is in fact an equality. We defer the proof, which relies on standard properties and a direct calculation of the variance of $\hat T$ around its mean $T$, to \Cref{sec:kernel-integral-operators}. In the following, we shall write $c = \kappa \, \normm{C}_\infty \, \expectt[x]{\norm{x}^2}$ to simplify calculations for the proof of \Cref{th:convergence_deep}, where $C = \expectt[w \sim \pi]{ww\trans}$ is the uncentered covariance of $\pi$, and $\kappa$ is a constant. When $\pi$ is Gaussian, \Cref{sec:kernel-integral-operators} further shows that $\kappa \leq \sqrt{3}$.

The mean-square error between $\hat\varphi$ and $\varphi$ after alignment can then be expressed as a different distance between $\hat T$ and $T$, as proven in the next lemma.
\begin{lemma}
    \label{th:bures-wasserstein}
    The alignment error between $\hat\varphi$ and $\varphi$ is equal to the Bures-Wasserstein distance $\bw$ between $\hat T$ and $T$:
    \begin{align*}
        \min_{\hat A \in \mathcal{O}\paren{d_1}} \expect[x]{ \normm{\hat A \, \hat \varphi(x) - \varphi(x)}^2_{H} } &= \bw\parenn{\hat T, T}^2.
    \end{align*}
\end{lemma}

The Bures-Wasserstein distance \citep{bhatia-bures-wasserstein} is defined, for any trace-class non-negative self-adjoint operators $\hat T$ and $T$, as
\begin{align*}
    \bw(\hat T, T)^2 = \min_{\hat A \in O\paren{\Ltwo}} \normm{\hat A \, \hat T\psqrtt - T\psqrtt}{}_2^2 = \tr\paren{\hat T + T - 2\paren{T\psqrtt \hat T T\psqrtt}\psqrtt}.
\end{align*}
The minimization in the first term is done over unitary operators of $\Ltwo$, and can be solved in closed-form with a singular value decomposition of $T\psqrtt \hat T\psqrtt$ as in \cref{eq:alignment_objective,eq:alignment_closed_form}. A direct calculation then shows that the minimal value is equal to the expression in the second term, as in \cref{eq:similarity_measure}. The Bures-Wasserstein distance arises in optimal transport as the Wasserstein-$2$ distance between two zero-mean Gaussian distributions of respective covariance operators $\hat T$ and $T$, and in quantum information as the Bures distance, a non-commutative generalization of the Hellinger distance. We refer the interested reader to \citet{bhatia-bures-wasserstein} for more details. We defer the proof of \Cref{th:bures-wasserstein} to \Cref{sec:proof-bures-wasserstein}.

It remains to establish the convergence of $\hat T$ towards $T$ for the Bures-Wasserstein distance, which is a distance on the square roots of the operators. The main difficulty comes from the fact that the square root is Lipschitz continuous only when bounded away from zero. This lack of regularity in the optimization problem can be seen from the fact that the optimal alignment rotation $\hat A$ is obtained by setting all singular values of some operator to one, which is unstable when this operator has vanishing singular values. We thus consider an entropic regularization of the underlying optimal transport problem over $\hat A$ with a parameter $\lambda > 0$ that will be adjusted with $d_1$. It penalizes the entropy of the coupling so that singular values smaller than $\lambda$ are not amplified. It leads to a bound on the Bures-Wasserstein distance, as shown in the following lemma.

\begin{lemma}
    \label{th:bures-wasserstein-regularization}
    Let $\hat T$ and $T$ be two trace-class non-negative self-adjoint operators. For any $\lambda > 0$, we have
    \begin{align}
        \label{eq:bw-control}
        \bw(\hat T, T)^2 \leq \frac{\norm{T}_2\normm{\hat T - T}_2}{\lambda} + \tr\parenn{\hat T - T} + 2\tr\paren{T + \lambda\Id - \paren{T^2 + \lambda^2 \Id}\psqrtt}.
    \end{align}
\end{lemma}
We defer the proof to \Cref{sec:proof-bures-wasserstein-regularization}.

The first two terms in \cref{eq:bw-control} are controlled in expectation with \Cref{th:kernel_mean_square_convergence}. The last term, when divided by $\lambda$, has a similar behavior to another quantity which arises in least-squares regression, namely the degrees of freedom $\tr\parenn{ T\paren{T + \lambda\Id}\pinv}$ \citep{hastie-tibshirani-generalized-additive-models,caponnetto-de-vito-regularized-least-squares}. It can be calculated by assuming a decay rate for the eigenvalues of $T$, as done in the next lemma.
\begin{lemma}
    \label{th:regularization-term-computation}
    Let $T$ be a trace-class non-negative self-adjoint operator whose eigenvalues satisfy $\lambda_m \leq c \, m^{-\alpha}$ for some $\alpha > 1$ and $c > 0$. Then it holds:
    \begin{align*}
        \tr\paren{T + \lambda \Id - \paren{T^2 + \lambda^2 \Id}\psqrtt} \leq  c' \, \lambda^{1 - 1/\alpha},
    \end{align*}
    where the constant $c' = \frac{c^{1/\alpha}}{1 - 1/\alpha}$.
\end{lemma}
The proof is in \Cref{sec:proof-regularization-term-computation}.

We now put together \Cref{th:kernel_mean_square_convergence,th:bures-wasserstein,th:bures-wasserstein-regularization,th:regularization-term-computation}. We have for any $\lambda > 0$,
\begin{align*}
    \expect[W,x]{\normm{\hat A \, \hat \varphi(x) - \varphi(x)}{}^2_{H}} &= \expect[W]{\bw(\hat T, T)^2} \leq \frac{\kappa \, \, \norm{C}_\infty^2 \expectt[x]{\norm{x}^2}^2}{\lambda \sqrt{d_1}} + \frac{2c^{1/\alpha}}{1 - 1/\alpha}  \lambda^{1-\frac1\alpha},
\end{align*}
where we have used the Cauchy-Schwarz inequality to bound $\expectt[W]{\normm{\hat T - T}_2} \leq \sqrt{\expectt[W]{\normm{\hat T - T}{}_2^2}}$ and the fact that $\norm{T}_2 \leq \tr T \leq \norm{C}_\infty \, \expectt[x]{\norm{x}^2}$. We then optimize the upper bound with respect to $\lambda$ by setting
\begin{align*}
    \lambda = \paren{ \frac{2 c^{1/\alpha} \sqrt{d_1}}{\kappa \norm{C}_\infty^2 \expectt[x]{\norm{x}^2}^2} }^{-\alpha/\paren{2\alpha - 1}},
\end{align*}
which yields
\begin{align*}
    \expect[W,x]{\normm{\hat A \, \hat \varphi(x) - \varphi(x)}{}^2_{H}} \leq c'' \, d_1^{-\paren{\alpha - 1}/\paren{4\alpha - 2}},
\end{align*}
with a constant
\begin{align*}
    c'' = \frac{2\kappa^{(\alpha-1)/(2\alpha-1)}}{(\alpha - 1)/\paren{2\alpha - 1}}  \paren{\frac{c}{\norm{C}_\infty \expectt[x]{\norm{x}^2}}}^{1/(2\alpha-1)} \norm{C}_\infty \, \expectt[x]{\norm{x}^2}.
\end{align*}

Finally, the function $\hat f$ can be written
\begin{align*}
    \hat f(x) = \innerr{\hat A\trans \theta, \hat\varphi(x)} = \innerr{\theta, \hat A \, \hat \varphi(x)}_H,
\end{align*}
so that
\begin{align*}
    \abss{\hat f(x) - f(x)}{}^2 = \abss{\innerr{\theta, \hat A \, \hat \varphi(x) - \varphi(x)}_H}{}^2 \leq \norm{\theta}^2_H \normm{\hat A \, \hat \varphi(x) - \varphi(x)}^2_{H}.
\end{align*}
Rewriting $\normm{\theta}_H = \normm{f}_\hilbert$, assuming that $\theta$ is the minimum-norm vector such that $f(x) = \inner{\theta, \varphi(x)}_H$, and using the convergence of $\hat A\, \hat \varphi$ towards $\varphi$ then yields
\begin{align*}
    \expect[W,x]{\abss{\hat f(x) - f(x)}{}^2} \leq c'' \norm{f}^2_\hilbert \, d_1^{-\paren{\alpha - 1}/\paren{4\alpha - 2}}.
\end{align*}

\subsection{Proof of \protect\Cref{th:kernel_mean_square_convergence}}
\label{sec:kernel-integral-operators}

We define the linear operator $\Phi \colon \Ltwo \to H$ by
\begin{align*}
    \Phi f = \expect[x]{f(x) \, \varphi(x)}.
\end{align*}
Its adjoint $\Phi\trans \colon H \to \Ltwo$ is then given by
\begin{align*}
    \parenn{\Phi\trans u}(x) = \inner{u, \varphi(x)},
\end{align*}
so that $T = \Phi\trans \Phi$. This proves that $T$ is self-adjoint and non-negative. On the other hand, we have $\Phi\Phi\trans = \expectt[x]{\varphi(x) \, \varphi(x)\trans}$ the uncentered covariance matrix of the feature map $\varphi$ associated to the kernel $k$. This shows that $T$ and this uncentered covariance matrix have the same eigenvalues.

Moreover, we have
\begin{align*}
    \tr(T) = \expect[x]{k(x,x)} = \tr\paren{\Phi\trans \Phi} = \norm{\Phi}_2^2 = \expect[x]{\normm{\varphi(x)}^2},
\end{align*}
and using the definition of $k$,
\begin{align*}
    \expect[x]{k(x,x)} = \expect[x,w]{\sigma\paren{\inner{w, x}}^2} \leq \expect[x,w]{\abs{\inner{w, x}}^2} = \tr \paren{C \, \expect[x]{xx\trans}} \leq \norm{C}_\infty \, \expect[x]{\norm{x}^2},
\end{align*}
where $w \sim \pi$ independently from $x$, $\abs{\sigma(t)} \leq \abs{t}$ by assumption on $\sigma$, and the last step follows from Hölder's inequality. This proves that $T$ is trace-class and $\Phi$ is Hilbert-Schmidt, with an explicit upper bound on the trace.

The above remarks are also valid for $\hat T$ with an appropriate definition of $\hat \Phi \colon \Ltwo \to \RR^{d_1}$. We have $\expectt[W]{\hat T} = T$ because $\expectt[W]{\hat k(x,x')} = k(x,x')$. Therefore, $\tr (\hat T) = \normm{\hat \Phi}{}_2^2$ is almost surely finite because
\begin{align*}
    \expectt[W]{\tr\parenn{\hat T}} = \tr(T) < + \infty.
\end{align*}

Let $\hat k_i(x,x') = \sigma\paren{\inner{w_i,x}} \, \sigma\parenn{\innerr{w_i,x'}}$ where $(w_i)_{i \leq d_j}$ are the rows of $W$, and $\hat T_i$ the associated integral operators. The $\hat T_i$ are i.i.d.\ with $\expectt[W]{\hat T_i} = T$ as for $\hat T$, and we have $\hat T = d_1\pinv \summ i{d_1} \hat T_i$. It then follows by standard variance calculations that
\begin{align*}
    \expect[W]{\normm{\hat T - T}{}_2^2}
    &= \frac1{d_1} \paren{\expect[W]{\normm{\hat T_1}{}_2^2} - \norm{T}_2^2} = \frac{c}{d_1},
\end{align*}
with a constant $c$ such that
\begin{align*}
    c \leq \expect[W]{\normm{\hat T_1}{}_2^2}
    \leq \expect[W]{\tr\parenn{\hat T_1}^2}
    = \expect[W]{ \expect[x]{\sigma\parenn{\inner{w_1,x}}^2 }^2 }
    \leq \expect[W]{ \expect[x]{\abss{\inner{w_1,x}}^2 }^2 }.
\end{align*}
We then have, using the assumption on the fourth moments of $\pi$,
\begin{align*}
    \expect[W]{ \expect[x]{\abss{\inner{w_1,x}}^2 }^2 }
    &= \expect[W]{ \paren{w_1\trans \expect[x]{xx\trans} w_1}^2 } < +\infty,
\end{align*}
because $\tr \expectt[x]{xx\trans} = \expectt[x]{\norm{x}^2} < +\infty$. When $\pi$ is Gaussian, we further have
\begin{align*}
    \expect[W]{ \expect[x]{\abss{\inner{w_1,x}}^2 }^2 } &= \paren{\tr\paren{C \, \expect[x]{xx\trans}}}^2 + 2 \tr\paren{\paren{C \, \expect[x]{xx\trans}}^2} \\
    &\leq 3\paren{\tr\paren{C \, \expect[x]{xx\trans}}}^2 \\
    &\leq 3 \, \norm{C}_\infty^2 \, \expect[x]{\norm{x}^2}^2,
\end{align*}
by classical fourth-moment computations of Gaussian random variables.

\subsection{Proof of \protect\Cref{th:bures-wasserstein}}
\label{sec:proof-bures-wasserstein}

The alignment error can be rewritten in terms of the linear operators $\Phi$ and $\hat \Phi$ defined in \Cref{sec:kernel-integral-operators}:
\begin{align*}
    \expect[x]{\normm{\hat A \, \hat\varphi(x) - \varphi(x)}{}_{H}^2} = \normm{\hat A \, \hat\Phi - \Phi}{}_2^2.
\end{align*}
We then expand
\begin{align*}
    \normm{\hat A \, \hat\Phi - \Phi}{}_2^2
    &= \normm{\hat\Phi}{}_2^2 + \normm{\Phi}_2^2 - 2\tr\paren{\Phi\trans \hat A\hat\Phi}.
\end{align*}

The first two terms are respectively equal to $\tr \hat T$ and $\tr T$ per \Cref{sec:kernel-integral-operators}. The alignment error is minimized with $\hat A = UV\trans$ from the SVD decomposition \citep{bhatia-bures-wasserstein}:
\begin{align*}
    \Phi \hat \Phi\trans = \expect[x]{\varphi(x) \, \hat\varphi(x)\trans} = USV\trans,
\end{align*}
for which we then have
\begin{align*}
    \tr\paren{\Phi\trans \hat A \hat\Phi} = \tr\paren{\hat\Phi \Phi\trans \hat A} = \tr\paren{VSU\trans UV\trans} = \tr\paren{S}.
\end{align*}

This can further be written
\begin{align*}
   \tr(S) = \tr\paren{\paren{U S^2 U\trans}\psqrtt} = \tr\paren{\paren{\Phi \hat\Phi\trans \hat\Phi \Phi\trans}\psqrtt} = \tr\paren{\paren{\Phi \hat T \Phi\trans}\psqrtt}.
\end{align*}
To rewrite this in terms of $T$, we perform a polar decomposition of $\Phi$: there exists a unitary operator $P \colon \Ltwo \to H$ such that $\Phi = PT\psqrtt$. We then have
\begin{align*}
\tr\paren{\paren{\Phi \hat T \Phi\trans}\psqrtt}
&= \tr\paren{\paren{P T\psqrtt \hat T T\psqrtt P\trans}\psqrtt} \\
&= \tr\paren{P \paren{T\psqrtt \hat T T\psqrtt}\psqrtt P\trans} \\
&= \tr\paren{\paren{T\psqrtt \hat T T\psqrtt}\psqrtt}.
\end{align*}

Putting everything together, we have
\begin{align*}
    \expect[x]{\normm{\hat A \, \hat\varphi(x) - \varphi(x)}{}_{H}^2} = \tr\paren{\hat T + T - 2 \paren{T\psqrtt \hat T T\psqrtt}\psqrtt}.
\end{align*}

\subsection{Proof of \protect\Cref{th:bures-wasserstein-regularization}}
\label{sec:proof-bures-wasserstein-regularization}

The Bures-Wasserstein distance can be rewritten as a minimum over contractions rather than unitary operators:
\begin{align*}
    \bw(\hat T, T)^2 = \min_{\normm{\hat A}_\infty \leq 1} \tr\paren{\hat T + T - 2 T\psqrtt \hat A \hat T\psqrtt},
\end{align*}
which holds because of Hölder's inequality:
\begin{align*}
    \tr\paren{T\psqrtt \hat A \hat T\psqrtt} = \tr\paren{\hat T\psqrtt T\psqrtt \hat A} \leq \normm{\hat T\psqrtt T\psqrtt}_1 \normm{\hat A}_\infty = \tr\paren{\paren{T\psqrtt \hat T T\psqrtt}\psqrtt} \normm{\hat A}_\infty.
\end{align*}
Rather than optimizing over contractions $\hat A$, which leads to a unitary $\hat A$, we shall use a non-unitary $\hat A$ with $\normm{\hat A}_\infty < 1$.

We introduce an ``entropic'' regularization: let $\lambda > 0$, and define
\begin{align*}
    \bw_\lambda(\hat T, T)^2 = \min_{\normm{\hat A}_\infty \leq 1} \tr\paren{\hat T + T - 2 T\psqrtt \hat A \hat T\psqrtt} + \lambda\logdet\paren{\paren{\Id - \hat A\trans \hat A}\pinv}.
\end{align*}
The second term corresponds to the negentropy of the coupling in the underlying optimal transport formulation of the Bures-Wasserstein distance.
It can be minimized in closed-form by calculating the fixed-point of Sinkhorn iterations \citep{peyre-entropic-transport-gaussians}, or with a direct SVD calculation as in \Cref{sec:proof-bures-wasserstein}. It is indeed clear that the minimum is attained at some $\hat A_\lambda = US_\lambda V\trans$ with $T\psqrtt \hat T\psqrtt = USV\trans$, and this becomes a separable quadratic problem over the singular values $S_\lambda$. We thus find
\begin{align*}
    S_\lambda &= \paren{\paren{S^2 + \lambda^2\Id}\psqrtt - \lambda\Id} S\pinv, \\
    \hat A_\lambda &= \paren{\paren{T\psqrtt \hat T T\psqrtt + \lambda^2 \Id}\psqrtt - \lambda \Id} T\prsqrt \hat T\prsqrt ,
\end{align*}
and one can verify that we indeed have $\normm{\hat A_\lambda}{}_\infty < 1$. When plugged in the original distance, it gives the following upper bound:
\begin{align*}
    \bw(\hat T, T)^2 \leq \tr\paren{\hat T + T - 2\paren{\paren{T\psqrtt \hat T T\psqrtt + \lambda^2\Id}\psqrtt - \lambda\Id}}.
\end{align*}

The term $\lambda^2 \Id$ in the square root makes this a Lipschitz continuous function of $\hat T$. Indeed, define the function $g$ by
\begin{align*}
    g(\hat T) &= \tr\paren{\paren{T\psqrtt \hat T T\psqrtt + \lambda^2\Id}\psqrtt - \lambda\Id}.
\end{align*}
Standard calculations \citep{bhatia-bures-wasserstein,peyre-entropic-transport-gaussians} then show that
\begin{align*}
    \nabla g(\hat T) &= \frac12 T\psqrtt \paren{T\psqrtt \hat T T\psqrtt + \lambda^2\Id}\prsqrtt T\psqrtt.
\end{align*}
It implies that
\begin{align*}
    0 \preccurlyeq \nabla g(\hat T) \preccurlyeq \frac1{2\lambda} T,
\end{align*}
where have used that $T\psqrtt \hat T T\psqrtt \succcurlyeq 0$ in the second inequality, and finally,
\begin{align*}
    \normm{\nabla g(\hat T)}_2 \leq \frac{\norm{T}_2}{2\lambda}.
\end{align*}
This last inequality follows from
\begin{align*}
    \normm{\nabla g(\hat T)}{}_2^2 = \tr\paren{\nabla g(\hat T) \trans \nabla g(\hat T)} \leq \tr\paren{\nabla g(\hat T)\trans \frac1{2\lambda}T} \leq \normm{\nabla g(\hat T)}_2 \frac{\normm{T}_2}{2\lambda},
\end{align*}
where we have used the operator-monotonicity of the map $M \mapsto \tr\paren{\nabla g(\hat T)\trans M}$, which holds because $\nabla g(\hat T) \succcurlyeq 0$.

Using the bound on the Lipschitz constant of $g$, we can then write
\begin{align*}
    \abss{g(\hat T) - g(T)}
    &\leq \frac{\norm{T}_2}{2\lambda} \normm{\hat T - T}_2.
\end{align*}
This leads to an inequality on the Bures-Wasserstein distance:
\begin{align*}
    \bw(\hat T, T)^2 &\leq \tr\parenn{\hat T + T} - 2g(\hat T) \\
    &= 2\paren{\tr(T) - g(T)} + \tr\parenn{\hat T - T} - 2\parenn{g(\hat T) - g(T)} \\
    &\leq 2\paren{\tr(T) - g(T)} + \tr\parenn{\hat T - T} + \frac{\norm{T}_2}{\lambda} \normm{\hat T-T}_2,
\end{align*}
which concludes the proof.

\subsection{Proof of \protect\Cref{th:regularization-term-computation}}
\label{sec:proof-regularization-term-computation}

We have
\begin{align*}
    \tr\paren{T + \lambda \Id - \paren{T^2 + \lambda^2 \Id}\psqrtt}
    &= \summ m\infty \paren{\lambda_m + \lambda - \sqrt{\lambda_m^2 + \lambda^2}}
\end{align*}
We have the following inequality
\begin{align*}
    {\lambda_m + \lambda - \sqrt{\lambda_m^2 + \lambda^2}} \leq \min\paren{\lambda_m, \lambda},
\end{align*}
by using $\sqrt{\lambda_m^2 + \lambda^2} \geq \max\paren{\lambda_m, \lambda}$.

We have $\lambda_m \leq c\, m^{-\alpha}$ for all $m$. We split the sum at $M = \lfloor \paren{\lambda / c}^{-1/\alpha} \rfloor$ (so that $c\, M^{-\alpha} \approx \lambda$), and we have
\begin{align*}
    \summ mM \paren{\lambda_m + \lambda - \sqrt{\lambda_m^2 + \lambda^2}} &\leq \summ mM \lambda = M\lambda, \\
    \sum_{m=M+1}^{\infty} \paren{\lambda_m + \lambda - \sqrt{\lambda_m^2 + \lambda^2}} &\leq \sum_{m=M+1}^{\infty} \lambda_m \leq c \sum_{m=M+1}^{\infty} m^{-\alpha} \leq c\frac{M^{1-\alpha}}{\alpha - 1},
\end{align*}
Finally,
\begin{align*}
    \summ 1\infty \paren{\lambda_m + \lambda - \sqrt{\lambda_m^2 + \lambda^2}} &\leq \paren{\frac{\lambda}c}^{-1/\alpha} \lambda + \frac{c}{\alpha - 1} \paren{\frac\lambda c}^{1 - 1/\alpha} = \frac{c^{1/\alpha}}{1 - 1/\alpha} \, \lambda^{1 - 1/\alpha}.
\end{align*}

\section{Proof of \protect\Cref{th:convergence_deep}}
\label{app:proof-convergence-deep}

In this section, expectations are taken with respect to both the weights $W_1, \dots, W_j$ and the input $x$. We remind that $W_j = W'_j \, \hat A_{j-1}$ with $W'_j$ having i.i.d.\ rows $w'_{ji} \sim \pi_j$. Let $C_j = \expectt[w_j\sim\pi_j]{w_jw_j\trans}$ be the uncentered covariance of $\pi_j$. Similarly to \Cref{app:proof-convergence-shallow}, we assume without loss of generality that $\sigma$ is $1$-Lipschitz and that $\sigma(0) = 0$.

Let $\tilde \phi_j = \sigma W'_j \, \phi_{j-1}$. Let $A_j \in \mathcal{O}(d_j)$ to be adjusted later. We have by definition of $\hat A_j$:
\begin{align}
    \sqrt{\expect{\normm{ \hat A_j \, \hat\phi_j(x) - \phi_j(x)}{}^2}}
    &\leq \sqrt{\expect{\normm{ A_j \, \hat\phi_j(x) - \phi_j(x)}{}^2}} \nonumber \\
    &\leq \sqrt{\expect{\normm{ A_j \hat\phi_{j}(x) - A_j \tilde\phi_{j}(x) }{}^2}} + \sqrt{\expect{\normm{A_j \tilde\phi_{j}(x) - \phi_j(x)}{}^2}},
    \label{eq:activation_error_decomp}
\end{align}
where the last step follows by the triangle inequality. We now bound separately each term.

To bound the first term, we compute the Lipschitz constant of $\sigma W'_j$ (in expectation). For any $z$, $z' \in H_{j-1}$, we have:
\begin{align*}
    \expect{\normm{{\sigma W'_j} z - {\sigma W'_j} z'}^2}
    &\leq \frac1{d_j} \expect{\normm{W'_j \parenn{z - z'}}^2} \\
    &= \frac1{d_j}\summ i{d_j} \expect{ \abss{\innerr{w'_{ji}, z - z'}}^2} \\
    &= \parenn{z-z'}\trans C_j \parenn{z-z'} \\
    &\leq \normm{C_j}{}_\infty \normm{z - z'}{}^2,
\end{align*}
where we have used the fact that $\sigma$ is $1$-Lipschitz, and have made explicit the normalization factor of $d_j\pinv$. We can therefore bound the first term in \cref{eq:activation_error_decomp}:
\begin{align*}
    \sqrt{\expect{\normm{ A_j \hat\phi_{j}(x) - A_j \tilde \phi_{j}(x) }{}^2}}
    &= \sqrt{\expect{\normm{ \parenn{\sigma W'_j} \hat A_{j-1} \hat\phi_{j-1}(x) - \parenn{\sigma W'_j} \phi_{j-1}(x) }{}^2}} \\
    & \leq \normm{C_j}{}\psqrtt_\infty \sqrt{\expect{\normm{\hat A_{j-1} \hat\phi_{j-1}(x) - \phi_{j-1}(x)}{}^2}}.
\end{align*}

We define $A_j$, which was arbitrary, as the minimizer of the second term in \cref{eq:activation_error_decomp} over $\mathcal O(d_j)$. We can then apply \Cref{th:convergence_shallow} to $z = \phi_{j-1}(x)$. Indeed, $\expectt[z]{\varphi_j(z) \, \varphi_j(z)\trans} = \expectt[x]{ \phi_j(x) \, \phi_j(x)\trans}$ is trace-class with eigenvalues $\lambda_{j,m} = O(m^{-\alpha_j})$, and $\pi_j$ has bounded second-{} and fourth-order moments. Therefore, there exists a constant $c_j$ such that
\begin{align*}
    \sqrt{\expect{\normm{A_j \tilde\phi_{j}(x) - \phi_j(x)}{}^2}}
    &= \sqrt{\expect{\normm{A_j \, \sigma W'_j \, \phi_{j-1}(x) - \varphi_j\, \phi_{j-1}(x)}{}^2}} \\
    &\leq \normm{C_j}_\infty\psqrtt \sqrt{\expectt{\norm{\phi_{j-1}(x)}^2}} \, c_j \, d_j^{-\eta_j/2},
\end{align*}
with $\eta_j = \frac{\alpha_j - 1}{2(2\alpha_j - 1)}$. We have made explicit the factors $\normm{C_j}_\infty\psqrtt \sqrt{\expectt{\norm{\phi_{j-1}(x)}^2}}$ in the constant coming from \Cref{th:convergence_shallow} to simplify the expressions in the sequel.
We can further bound $\sqrt{\expectt{\normm{\phi_{j-1}(x)}^2}}$ by iteratively applying \Cref{th:kernel_mean_square_convergence} from \Cref{app:proof-convergence-shallow}:
\begin{align*}
    \sqrt{\expectt{\normm{\phi_{j-1}(x)}^2}} \leq \normm{C_{j-1}}{}_\infty\psqrtt \cdots \normm{C_1}{}_\infty\psqrtt \sqrt{\expectt{\norm{x}^2}}.
\end{align*}

We thus have shown:
\begin{align*}
    \sqrt{\expect{\normm{ \hat A_j \, \hat\phi_j(x) - \phi_j(x)}{}^2}}
    &\leq \normm{C_j}{}\psqrtt_\infty \sqrt{\expect{\normm{\hat A_{j-1} \hat\phi_{j-1}(x) - \phi_{j-1}(x)}{}^2}} \\
    & \quad + \normm{C_{j}}{}_\infty\psqrtt \cdots \normm{C_1}{}_\infty\psqrtt \sqrt{\expectt{\norm{x}^2}} \, c_j \, d_j^{-\eta_j/2}.
\end{align*}
It then follows by induction:
\begin{align*}
    \sqrt{\expect{\normm{ \hat A_j \, \hat\phi_j(x) - \phi_j(x)}{}^2}}
    &\leq \normm{C_{j}}{}_\infty\psqrtt \cdots \normm{C_1}{}_\infty\psqrtt \sqrt{\expectt{\norm{x}^2}} \, \summ \ell j c_\ell \, d_\ell^{-\eta_\ell/2}.
\end{align*}
We conclude like in the proof of \Cref{th:convergence_shallow}:
\begin{align*}
    \sqrt{\expect{\abss{ \hat f(x) - f(x)}{}^2}}
    &\leq \normm{f}_{\hilbert_J} \normm{C_J}{}_\infty\psqrtt \cdots \normm{C_{1}}{}_\infty\psqrtt \sqrt{\expectt{\norm{x}^2}} \, \summ jJ  c_j \, d_j^{-\eta_j/2}.
\end{align*}

We finally show the convergence of the kernels. Let $\tilde k_j$ be the kernel defined by the feature map $\tilde \phi_j$. Expectations are now also taken with respect to $x'$, an i.i.d.\ copy of $x$. We have by the triangle inequality:
\begin{align}
    \label{eq:kernel_decomposition}
    \abss{\hat k_j(x,x') - k_j(x,x')}
    &\leq \abss{\hat k_j(x,x') - \tilde k_j(x,x')} + \abss{\tilde k_j(x,x') - k_j(x,x')}.
\end{align}

For the first term on the right-hand side:
\begin{align*}
    \abss{\hat k_j(x,x') - \tilde k_j(x,x')}
    &= \abss{\innerr{\hat \phi_j(x), \hat\phi_j(x')} - \innerr{\tilde\phi_j(x), \tilde\phi_j(x')}} \\
    &\leq \abss{\innerr{\hat\phi_j(x), \hat\phi_j(x') - \tilde\phi_j(x')} + \innerr{\hat\phi_j(x) - \tilde\phi_j(x),\tilde\phi_j(x')}} \\
    &\leq \normm{\hat\phi_j(x)}\normm{\hat\phi_j(x') - \tilde\phi_j(x')} + \normm{\tilde\phi_j(x')}\normm{\hat\phi_j(x) - \tilde\phi_j(x)}.
\end{align*}
We thus have, because $x,x'$ are i.i.d.,
\begin{align*}
    &\sqrt{\expect{\abss{\hat k_j(x,x') - \tilde k_j(x,x')}{}^2}} \\
    &\leq \sqrt{\expect{\normm{\hat\phi_j(x)}{}^2} \expect{\normm{\hat\phi_j(x') - \tilde\phi_j(x')}{}^2}} + \sqrt{\expect{\normm{\tilde\phi_j(x')}{}^2} \expect{\normm{\hat\phi_j(x) - \tilde\phi_j(x)}{}^2}}.
\end{align*}
Using the Lipschitz constant of $\sigma W'_j$ in expectation as above:
\begin{align*}
    \sqrt{\expect{\abss{\hat k_j(x,x') - \tilde k_j(x,x')}{}^2}}
    &\leq 2 \normm{C_j}{}_\infty
    \sqrt{\expect{\normm{\phi_{j-1}(x)}{}^2} \expect{\normm{\hat\phi_{j-1}(x) - \phi_{j-1}(x)}{}^2}}.
\end{align*}
The factors on the right-hand side can be bounded using the above, to yield
\begin{align*}
    \sqrt{\expect{\abss{\hat k_j(x,x') - \tilde k_j(x,x')}{}^2}}
    &\leq 2 \normm{C_j}{}_\infty \cdots \normm{C_1}{}_\infty \expectt{\norm{x}^2} \, \summ \ell {j-1} c_\ell \, d_\ell^{-\eta_\ell/2}.
\end{align*}

The second term on the right-hand side of \cref{eq:kernel_decomposition} can be bounded with \Cref{th:convergence_shallow} applied to $z = \phi_{j-1}(x)$ as before:
\begin{align*}
    \sqrt{\expect{\abss{\tilde k_j(x,x') - k_j(x,x')}{}^2}} \leq \kappa_j \normm{C_j}{}_\infty \cdots \normm{C_1}{}_\infty {\expectt{\norm{x}^2}} \, d_j\prsqrtt.
\end{align*}
where we have again used the upper bound on $\sqrt{\expectt{\norm{\phi_{j-1}(x)}^2}}$.

We thus have shown that
\begin{align*}
    \sqrt{\expect{\abss{ k_j(x,x') - k_j(x,x')}{}^2}} \leq \normm{C_j}{}_\infty \cdots \normm{C_1}{}_\infty {\expectt{\norm{x}^2}} \, \paren{2\summ \ell {j-1} c_\ell \, d_\ell^{-\eta_\ell/2} + \kappa_j\, d_j\prsqrtt}.
\end{align*}

\section{Proof of \protect \Cref{th:rainbow_activations_equivariance}}
\label{app:proof_rainbow_activations_equivariance}

We prove the result by induction on the layer index $j$. We initialize with $\phi_0(x) = x$, which admits an orthogonal representation $\rho_0(g) = g$. Now suppose that $\phi_{j-1}$ admits an orthogonal representation $\rho_{j-1}$. Let $w \sim \pi_j$, we have that $\rho_{j-1}(g) \trans w \sim \pi_j$ for all $g \in G$ by hypothesis. When $\pi_j = \normal(0, C_j)$, this is equivalent to $\rho_{j-1}(g)\trans C_j \rho_{j-1}(g) = C_j$, i.e.\ $\rho_{j-1}(g) C_j = C_j \rho_{j-1}(g)$. We begin by showing that $\phi_j$ then admits an orthogonal representation $\rho_j$.

We have
\begin{align*}
    \phi_j(g x) = \varphi_j(\phi_{j-1}(gx)) = \varphi_j(\rho_{j-1}(g) \phi_{j-1}(x)).
\end{align*}
For simplicity, here we define the feature map $\varphi_j$ with $\varphi_j(z)(w) = \sigma(\inner{w, z})$ with $H_j = L^2(\pi_j)$ (the result of the theorem does however not depend on this choice, as all feature maps are related by a rotation). Then,
\begin{align*}
    \phi_j(g x)(w) = \sigma\paren{\inner{w, \rho_{j-1}(g) \phi_{j-1}(x)}} = \sigma\parenn{\innerr{\rho_{j-1}(g)\trans w, \phi_{j-1}(x)}}.
\end{align*}

For each $g \in G$, we thus define the operator $\rho_j(g)$ by its action on $\psi \in H_j$:
\begin{align*}
    (\rho_j(g) \psi)(w) = \psi(\rho_{j-1}(g)\trans w).
\end{align*}
It is obviously linear, and bounded as $\norm{\rho_j(g)}_\infty = 1$:
\begin{align*}
    \norm{\rho_j(g) \psi}^2_{H_j} = \expect[w]{\psi(\rho_{j-1}(g)\trans w)^2} = \expect[w]{\psi(w)^2} = \norm{\psi}^2_{H_j},
\end{align*}
where we have used that $\rho_{j-1}(g)\trans w \sim w$.
We further verify that $\rho_j(gg') = \rho_j(g)\rho_j(g')$:
\begin{align*}
    (\rho_j(gg')\psi)(w) &= \psi(\rho_{j-1}(gg')\trans w) = \psi(\rho_{j-1}(g')\trans \rho_{j-1}(g)\trans w) \\
    &= (\rho_j(g')\psi)(\rho_{j-1}(g)\trans w) = (\rho_j(g) \rho_j(g') \psi)(w).
\end{align*}
We can thus write $\phi_j(gx) = \rho_j(g) \phi_j(x)$, which shows that $\phi_j$ admits a representation.

It remains to show that $\rho_j(g)$ is orthogonal. The adjoint $\rho_j(g)\trans$ is equal to $\rho_j(g\trans)$:
\begin{align*}
    \inner{\rho_j(g)\psi, \psi'}_{H_j} = \expect[w]{ \psi(\rho_{j-1}(g)\trans w) \psi'(w)} = \expect[w]{ \psi(w) \psi'(\rho_{j-1}(g) w)} = \inner{\psi, \rho_j(g\trans) \psi'}_{H_j},
\end{align*}
where we have used $\rho_{j-1}(g)\trans = \rho_{j-1}(g\trans)$ since $\rho_{j-1}$ is a group homomorphism. It is then straightforward that $\rho_j(g)\rho_j(g)\trans = \rho_j(g)\trans \rho_j(g) = \Id$ by using again the fact that $\rho_j$ is a group homomorphism. This proves that $\rho_j(g) \in {O}(H_{j})$.

We finally show that the rainbow kernel $k_j$ is invariant. We have
\begin{align*}
    k_j(gx, gx') &= \inner{\phi_j(gx), \phi_j(gx')}_{H_j} = \inner{\rho_j(g)\phi_j(x), \rho_j(g)\phi_j(x')}_{H_j} \\
    &= \inner{\phi_j(x), \phi_j(x')}_{H_j} = k_j(x, x'),
\end{align*}
which concludes the proof.

\section{Experimental details}
\label{app:architectures}

\paragraph{Normalization.}
In all the networks considered in this paper, after each non-linearity $\sigma$, a 2D batch-normalization layer \citep{batchnorm} without learned affine parameters sets the per-channel mean and variance across space and data samples to $0$ and $1$ respectively. After training, we multiply the learned standard deviations by $1/\sqrt{d_j}$ and the learned weight matrices $L_{j+1}$ by $\sqrt{d_j}$ as per our normalization conventions. This ensures that $\expectt[x]{\hat\phi_j(x)} = 0$ and $\expectt[x]{\normm{\hat\phi_j(x)}{}^2} = 1$, which enables more direct comparisons between networks of different sizes. When evaluating activation convergence for ResNet-18, we explicitly compute these expectations on the training set and standardize the activations $\hat\phi_j(x)$ after training for additional numerical stability. When sampling weights from the Gaussian rainbow model, the mean and variance parameters of the normalization layers are computed on the training set before alignment and sampling of the next layer.

\paragraph{Scattering networks.}
We use the learned scattering architecture of \citet{phase-collapse}, with several simplifications based on the setting.

The prior operator $P_j$ performs a convolution of every channel of its input with predefined filters: one real low-pass Gabor filter $\phi$ (a Gaussian window) and $4$ oriented Morlet wavelets $\psi_\theta$ (complex exponentials localized with a Gaussian window) \citep{kymatio}. $P_j$ also implements a subsampling by a factor $2$ on even layer indices $j$, with a slight modification of the filters to compute wavelet coefficients at intermediate scales. See \citet[Appendix G]{phase-collapse} for a precise definition of the filters. The learned weight matrices $L_j$ are real for CIFAR-10 experiments, and complex for ImageNet experiments.

We impose a commutation property between $P_j$ and $L_j$, so that we implement $W_j = P_j \, L_j$. It is equivalent to having $W_j = L_j \, P_j$, with the constraint that $L_j$ is applied pointwise with respect to the channels created by $P_j$. The non-linearity $\sigma$ is a complex modulus, which is only applied on the high-frequency channels. A scattering layer writes:
\begin{align*}
    \sigma W_j z = \paren{ L_j z \ast \phi, \abs{L_j z \ast \psi_\theta}}_{\theta}.
\end{align*}
The input (and therefore output) of $L_j$ are then both real when $L_j$ is real.

We apply a pre-processing $\sigma P_0$ to the input $x$ before feeding it to the network. The fully-connected classifier $\theta$ is preceded with a learned $1 \times 1$ convolution $L_{J+1}$ which reduces the channel dimension. The learned scattering architecture thus writes:
\begin{align*}
    \hat f(x) = \theta\trans L_{J+1} \, \sigma P_J L_J \, \cdots \, \sigma P_1 L_1 \, \sigma P_0 x.
\end{align*}
The number of output channels of $L_j$ is given in \Cref{tab:widths}.

As explained above, we include a 2D batch-normalization layer without learned affine parameters after each non-linearity $\sigma$, as well as before the classifier $\theta$. Furthermore, after each operator $L_j$, a divisive normalization sets the norm along channels at each spatial location to $1$ (except in \Cref{fig:covariance_convergence,fig:spectra_multilayer,fig:sgd_dynamics}). There are no learned biases in the architecture beyond the unsupervised channel means.

The non-linearity $\sigma$ includes a skip-connection in \Cref{fig:resampling,fig:spectra_multilayer}, in which case a scattering layer computes
\begin{align*}
    \sigma W_j z = \paren{ L_j z \ast \phi, L_j z \ast \psi_\theta, \abs{L_j z \ast \phi}, \abs{L_j z \ast \psi_\theta}}_{\theta}.
\end{align*}
In this case, the activations $\phi_j(x)$ are complex. The rainbow model extends to this case by adding complex conjugates at appropriate places. For instance, the alignment matrices become complex unitary operators when both activations and weights are complex.

\begin{table}[ht]
\small
\centering
\begin{tabular}{l r *{11}{c}}
\toprule
& $j$ & 1 & 2 & 3 & 4 & 5 & 6 & 7 & 8 & 9 & 10 & 11 \\
\midrule
\textbf{CIFAR-10 ($J = 3$)} & $d_j$ & 64 & 128 & 256 & 512 & - & - & - & - & - & - & - \\
\textbf{CIFAR-10 ($J = 7$)} & $d_j$ & 64 & 128 & 256 & 512 & 512 & 512 & 512 & 512 & - & - & - \\
\midrule
\textbf{ImageNet ($J = 10$)} & $d_j$ & 32 & 64 & 64 & 128 & 256 & 512 & 512 & 512 & 512 & 512 & 256 \\
\bottomrule \\
\end{tabular}
\caption{Number $d_j$ of output channels of $L_j$, $1 \leq j \leq J + 1$. The total number of projectors is $J + 1= 4$ or $J + 1 = 8$ for CIFAR-10 and $J + 1 = 11$ for ImageNet.}
\label{tab:widths}
\end{table}

\paragraph{ResNet.}
$P_j$ is the patch-extraction operator defined in \Cref{sec:prior_convolutional_architectures}. The non-linearity $\sigma$ is a ReLU. We have trained a slightly different ResNet with no bias parameters. In addition, the batch-normalization layers have no learned affine parameters, and are placed after the non-linearity to be consistent with our normalization conventions. The top-5 test accuracy on ImageNet remains at $89\%$ like the original model.

\paragraph{Training.}
Network weights are initialized with i.i.d.\ samples from an uniform distribution \citep{glorot-bengio-uniform-initialization} with so-called Kaiming variance scaling \citep{kaiming-he-initialization-variance}, which is the default in the PyTorch library \citep{pytorch}. Despite the uniform initialization, weight marginals become Gaussian after a single training epoch. Scattering networks are trained for $150$ epochs with an initial learning rate of $0.01$ which is divided by $10$ every $50$ epochs, with a batch size of $128$. ResNets are trained for $90$ epochs with an initial learning rate of $0.1$ which is divided by $10$ every $30$ epochs, with a batch size of $256$. We use the optimizer SGD with a momentum of $0.9$ and a weight decay of ${10}^{-4}$ (except for \Cref{fig:covariance_convergence,fig:sgd_dynamics} where weight decay has been disabled). We use classical data augmentations: horizontal flips and random crops for CIFAR, random resized crops of size $224$ and horizontal flips for ImageNet. The classification error on the ImageNet validation set is computed on a single center crop of size $224$.

\paragraph{Activation covariances.}
The covariance of the activations $\hat\phi_j(x)$ is computed over channels and averaged across space. Precisely, we compute
\begin{align*}
    \expect[x]{ \sum_u \hat\phi_j(x)[u] \, \hat\phi_j(x)[u]\trans },
\end{align*}
where $\hat\phi_{j}(x)[u]$ is a channel vector of dimension $d'_j$ at spatial location $u$. It yields a matrix of dimension $d'_{j} \times d'_{j}$. For scattering networks, the $d'_j$ channels correspond to the $d_{j}$ output channels of $L_j$ times the $5$ scattering channels computed by $P_j$ (times $2$ when $\sigma$ includes a skip-connection). For ResNet, $\hat\phi_j(x)[u]$ is a patch of size $s_j \times s_j$ centered at $u$ due to the operator $P_j$. $d_j$ is thus equal to the number $d'_j$ of channels of $\hat\phi_j$ multiplied by $s_j^2$.

\vskip 0.2in
\bibliography{refs}

\end{document}